# Improving Significant Wave Height Prediction Using Chronos Models

Yilin Zhai[1], Hongyuan Shi[1,2*], Chao Zhan[1,2], Qing Wang[1,2], Zaijin You[1,3], Nan Wang[3,4]

[1]School of Hydraulic and Civil Engineering, Ludong University, Shandong, Yantai, China

[2]Institute of Coastal Research, Ludong University, Shandong, Yantai, China

[3]Navigation College, Dalian Maritime University, Dalian, Liaoning, China

[4]Centre for Ports and Maritime Safety, Dalian Maritime University, Dalian, Liaoning, China

**\* Correspondence:**
Corresponding Author
hyshi@ldu.edu.cn

**Keywords: Ocean Wave prediction, Large language models (LLMs), Time-series forecasting, AutoML, Chronos.**

## Abstract

Accurate wave height prediction is critical for maritime safety and coastal resilience, yet conventional physics-based models and traditional machine learning methods face challenges in computational efficiency and nonlinear dynamics modeling. This study introduces Chronos, the first implementation of a large language model (LLM)-powered temporal architecture (Chronos) optimized for wave forecasting. Through advanced temporal pattern recognition applied to historical wave data from three strategically chosen marine zones in the Northwest Pacific basin, our framework achieves multimodal improvements: (1) 14.3% reduction in training time with 2.5× faster inference speed compared to PatchTST baselines, achieving 0.575 mean absolute scaled error (MASE) units; (2) superior short-term forecasting (1-24h) across comprehensive metrics; (3) sustained predictive leadership in extended-range forecasts (1-120h); and (4) demonstrated zero-shot capability maintaining median performance (rank 4/12) against specialized operational models. This LLM-enhanced temporal modeling paradigm establishes a new standard in wave prediction, offering both computationally efficient solutions and a transferable framework for complex geophysical systems modeling.

## 1    Introduction

Accurate ocean wave height forecasting is crucial for a wide range of maritime activities, including offshore operations, coastal engineering, and renewable energy generation (Kumar et al., 2017; Varela et al., 2024). For example, real-time wave height prediction enables proactive adjustments to the damping coefficient and power threshold of wave energy converters (WECs), thereby preventing device damage under extreme wave conditions (Zhang et al., 2022). Recent studies has demonstrated that this predictive technology can double the electricity generation output compared to conventional methods, although its effectiveness remains directly dependent on prediction accuracy (Li et al., 2012).

Over the past decades, substantial progress have been made in wave modeling, with the development of third-generation models like WAM (Group, 1988), SWAN (Booij et al., 1999; Liang et al., 2019) and WAVEWATCH III (Tolman, n.d.), which have significantly enhanced our understanding of wave dynamics (Gao et al., 2023). However, traditional physics-based models often struggle with computational complexity and require extensive parameter tuning, making them less scalable for real-time forecasting (Oh and Suh, 2018). This limitation motivates the exploration of alternative methods.

In response to these challenges, deep learning (DL) and machine learning (ML) approaches have emerged as promising alternatives for wave forecasting, offering significant advantages over traditional numerical and statistical methods. These data-driven techniques excel in capturing complex temporal patterns and nonlinear relationships inherent in wave data, enabling more accurate and adaptive forecasting (Yang et al., 2021; Gao et al., 2023). Unlike conventional physics-based models, which often require extensive computational resources and are sensitive to parameter tuning, DL and ML methods leverage large-scale historical datasets to learn intricate dependencies, making them particularly suitable for dynamic and highly variable marine environments (Reichstein et al., 2019).Various neural network architectures have been explored for wave height prediction, among which recurrent neural networks (RNNs) and their advanced variants, such as Long Short-Term Memory (LSTM) networks and Gated Recurrent Units (GRUs), have demonstrated notable success (Rumelhart et al., 1986; Cho et al., 2014; Kaur and Mohta, 2019; Zhang et al., 2024). For example, (Fan et al., 2020) developed an LSTM-based wave height prediction model trained on buoy-measured wave data from the National Data Buoy Center (NDBC), showing that LSTM outperformed traditional Support Vector Machine (SVM) models in both short-term and long-term wave height predictions. Similarly, (Alfredo and Adytia, 2022; Yevnin et al., 2023) applied a GRU-based framework to forecast significant wave heights in coastal regions, demonstrating its effectiveness in handling rapidly changing wave conditions while maintaining computational efficiency.

More recently, Transformer (Vaswani et al., 2023)-based models have gained increasing attention due to their self-attention mechanism, which enables them to model long-range dependencies more effectively than conventional RNN-based approaches (Karita et al., 2019). (Liu et al., 2024a; Wang et al., 2024) proposed a Transformer-based time-series prediction framework that has been adapted for ocean wave forecasting, showing its ability to maintain accuracy over extended prediction horizons. (Liu et al., 2024b) introduced a Spatiotemporal Transformer Network (STTN) for multi-step wave height forecasting, incorporating both temporal dependencies and spatial correlations from multiple buoy stations. Their results demonstrated that Transformer-based models outperform LSTM and GRU in long-term forecasting tasks, particularly in capturing extreme wave events and seasonal variations. Despite significant advancements, the use of Transformer-based models in wave forecasting remains insufficiently studied, particularly in marine environments with non-stationary and highly dynamic wave patterns. While previous studies have demonstrated their effectiveness in capturing long-term dependencies, their robustness and adaptability to varying sea states and extreme wave events warrant further investigation.

Unlike conventional neural network architectures that require extensive training on domain-specific datasets, Chronos (Ansari et al., 2024) integrates a pre-trained zero-shot learning mechanism, enabling it to generalize across diverse forecasting tasks with minimal fine-tuning. This ability to operate with limited task-specific data is particularly advantageous in wave forecasting, where rapidly changing marine conditions make it difficult to obtain comprehensive labeled datasets. Therefore, evaluating Chronos' performance in such environments is crucial for enhancing real-world operational wave forecasting.

To address these challenges, this study investigates the application of Chronos (Ansari et al., 2024) for wave height forecasting, systematically comparing its zero-shot performance against fine-tuned variants of Chronos and other state-of-the-art forecasting models (Pourpanah et al., 2023). Specifically, we evaluate 1–120 hour predictions using historical wave height data, aiming to address two key questions: (1) Can Chronos outperform traditional parametric models (e.g., AutoETS (Wu et al., 2021), DynamicOptimizedTheta (Fiorucci et al., 2016), SeasonalNative(Duong et al., 2023), NPTS), deep learning architectures (e.g., PatchTST (Nie et al., 2023; Huang et al., 2024), TiDE (Das et al., 2024), DeepAR (Salinas et al., 2019), Temporal Fusion Transformer (TFT)(Lim et al., 2020)) and tabular learning approaches (Gorishniy et al., 2023) (e.g., DirectTabular, RecursiveTabular (Erickson et al., 2020)) in marine settings? (2) Does domain-specific fine-tuning enhance Chronos' accuracy compared to its zero-shot baseline?

Our contributions are threefold: (1) First systematic evaluation of large language model (LLM)-based temporal modeling for ocean wave forecasting, extending Chronos' applications beyond typical time series domains. (2) Rigorous benchmarking against 10 established methods, revealing the advantages of attention mechanisms in capturing wave intermittency and seasonal trends. (3) Through systematic fine-tuning strategies applied to marine data, the ChronosFineTuned model demonstrates significant performance improvements over traditional recursive and non-parametric baselines, achieving reductions in MAE (1.1–37.45%), RMSE (0.89–34.78%), SMAPE (1.24–33.66%), RMSLE (0.8–34.57%) and MASE (0.96–35.38%).(4) This work bridges the gap between LLM-driven temporal modeling and ocean engineering, offering a scalable framework for operational wave forecasting.

The remainder of this paper is organized as follows. Chapter 2 presents the measured data and detailed information of five stations with different meteorological oceanographic conditions as well as the algorithmic architecture of the time-series model and the experimental design. In Chapter 3, numerical simulations of the proposed method are carried out using in situ observations and necessary discussions are made on the comparison of the accuracy of the different methods. Conclusions are summarized in Chapter 4.

## 2 Materials and methods

### 2.1 Data description

In this study, five buoy stations—41008, 41010, 42003, 44007, and 51003—were selected for short-term significant wave height (SWH) prediction modeling. These stations are strategically distributed across distinct marine regions, characterized by diverse water depths and environmental conditions. This diversity, spanning shallow coastal waters (e.g., 16 m at Station 41008) to deep ocean environments (e.g., 5023 m at Station 51003), enables a thorough evaluation of model adaptability across varying wave regimes.

The data were obtained from the National Oceanic and Atmospheric Administration (NOAA) buoy observation database, with observational records ranging from 15 to 33 years. This extensive temporal coverage captures a wide range of conditions, including extreme weather events such as typhoons and hurricanes, which are essential for developing and validating robust predictive models. Figure 1 illustrates the geographic distribution of the stations, providing a visual representation of their spatial coverage, while Table 1 details their coordinates, water depths, record spans, median and maximum SWH, and data volumes.

The primary objective of this study is to predict SWH time series at each station, with a focus on assessing the performance of the Chronos model against other forecasting methods. This comparative analysis evaluates the model's accuracy and adaptability under diverse regional wave conditions, which vary significantly as evidenced by median SWH values (0.92 m to 2.22 m) and maximum SWH values (5.84 m to 11.04 m). The large data volumes, ranging from 131,496 to 289,295 records per station, provide a robust foundation for model training and validation, ensuring reliable and generalizable results.

Table 1 summarizes the key characteristics of the selected stations, highlighting their observational timespans and data properties. The inclusion of long-term records and extreme events enhances the dataset's suitability for testing model performance in real-world scenarios, a critical consideration for applications in ocean engineering.

**Figure 1. Locations of Five NDBC Buoy Stations Across Diverse Marine Regions**

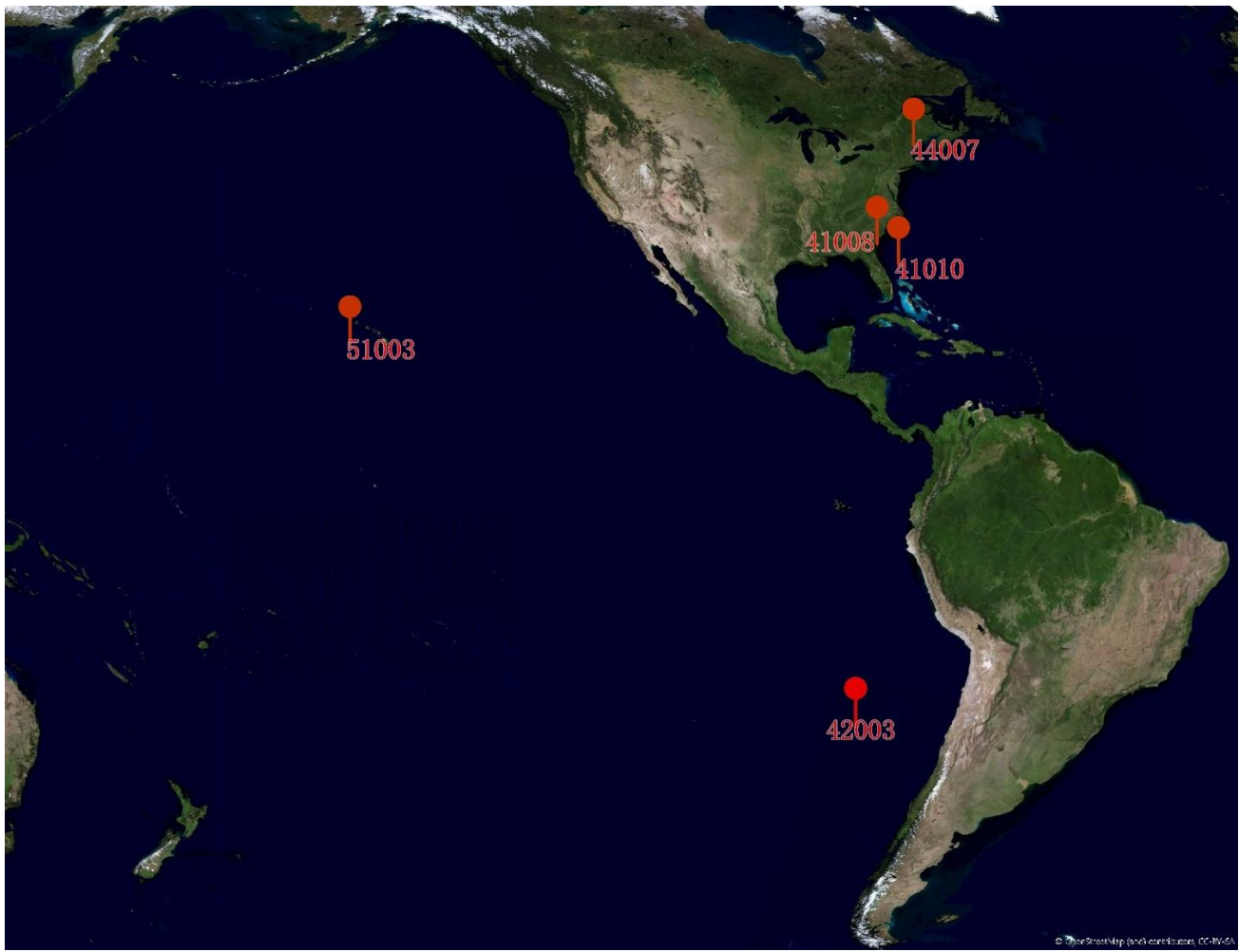

Table 1. Details of selected stations.

| Station ID | Coordinates | Depth (m) | Record Span | Median SWH (m) | Max SWH (m) | Data Volume |
|---|---|---|---|---|---|---|
| 41008 | 31.400°N, 80.866°W | 16 | Jan. 1, 1998 - Dec. 31, 2022 | 0.98 | 5.84 | 219144 |
| 41010 | 28.878°N, 78.467°W | 888 | Nov. 10, 1988 - Dec. 31, 2016 | 1.56 | 10.18 | 246678 |

| Station ID | Coordinates | Depth (m) | Record Span | Median SWH (m) | Max SWH (m) | Data Volume |
|---|---|---|---|---|---|---|
| 42003 | 25.925°N, 85.616°W | 3273 | Jan. 1, 2003 - Dec. 31, 2017 | 1.13 | 11.04 | 131496 |
| 44007 | 43.525°N, 70.140°W | 49 | Jan. 1, 1988 - Dec. 31, 2020 | 0.92 | 9.64 | 289295 |
| 51003 | 19.143°N, 160.645°W | 5023 | Apr. 29, 1988 - Dec. 31, 2018 | 2.22 | 6.85 | 268896 |

## 2.2 Chronos Model

The architecture of Chronos is shown in Figure 2. Chronos is a pre-trained probabilistic modeling framework for time series prediction. Its core idea is to scale and quantize the discretization of time series values into a fixed vocabulary, and to use cross-entropy loss to train existing transformer-based language model architectures on these discretized time series. Its pre-training was performed on many publicly available datasets using the T5 family (with parameters ranging from 20M to 710M), supplemented with a synthetic dataset generated by a Gaussian process to improve generalization.

**Figure 2 The architecture of Chronos**

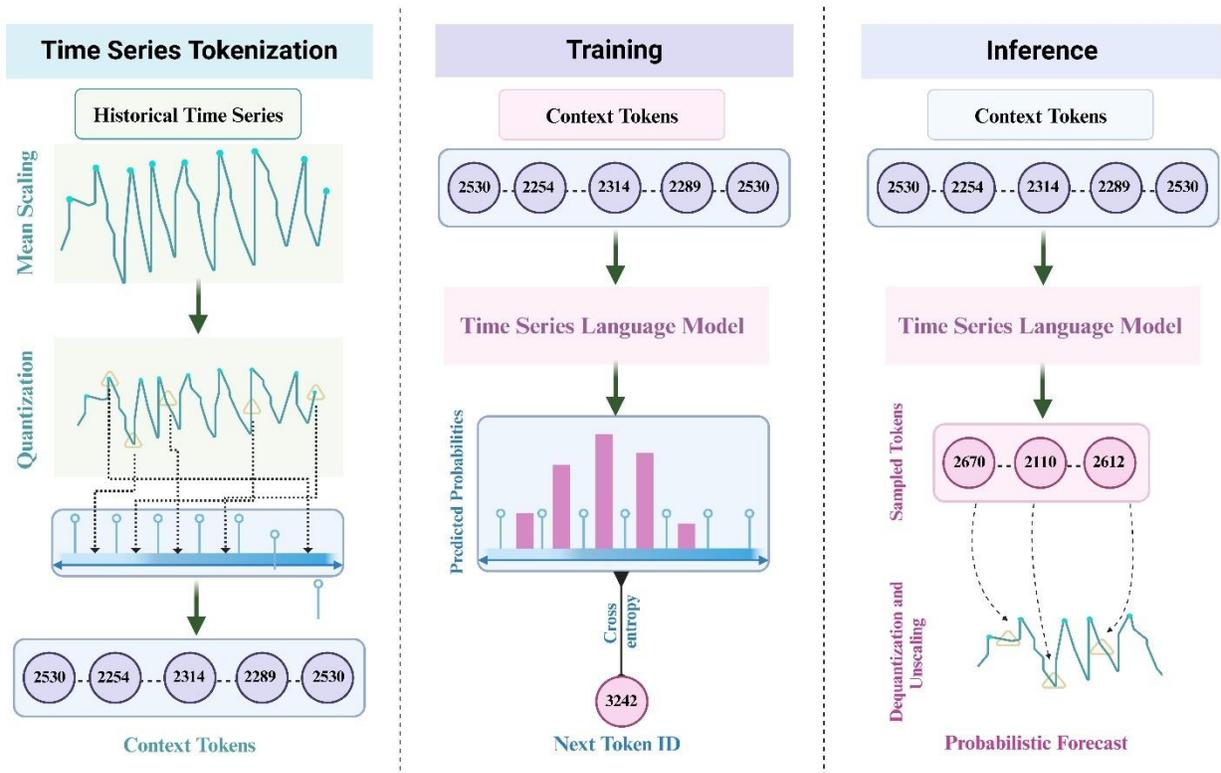

### 2.2.1 Time Series Tokenization

Consider a time series $x_{1:C+H} = [x_1, ..., x_{C+H}]$ where the first $C$ time steps constitute the historical context, and the remaining $H$ represent the forecast horizon. Language models operate on tokens from a finite vocabulary, so using them for time series data requires mapping the observations $x_i \in \mathbb{R}$ to a finite set of tokens.

**Scaling.** Mean scaling normalizes individual entries of the time series by the mean of the absolute values in the historical context. Specifically, this involves setting $m = 0$ and $s = \frac{1}{C}\sum i_1 = C|x_i|$.

**Quantization.** The scaled time series $\tilde{x}_{1:C+H} = [\tilde{x}_1, ..., \tilde{x}_C, ..., \tilde{x}_{C+H}]$, is still real-valued and cannot be processed directly by language models. To convert these real values into discrete tokens, we employ quantization. Formally, we select B bin centers $c_1 < \cdots < c_B$ on the real line, and $B - 1$ edges $b_i$ separating them, $c_i < b_i < c_{i+1}$, for $i \in \{1, \cdots, B-1\}$. The quantization function $q: \mathbb{R} \to \{1, 2, ..., B\}$, and dequantization $d: \{1, 2, ..., B\} \to \mathbb{R}$, are then defined as

$$q(x) = \begin{matrix} 1 & \text{if } -\infty \leq x < b_1 \\ 2 & \text{if } b_1 \leq x < b_2 \\ \cdots & \\ B & \text{if } b_{B-1} \leq x < \infty \end{matrix} \quad \text{and} \quad d(j) = c_j \quad (1)$$

respectively.

### 2.2.2 Objective Function

As typical in language models, we use the categorical distribution over the elements of $\mathcal{V}_{ts}$ as the output distribution, $p(z_{C+h+1}|z_{1:C+h})$ where $z_{1:C+h}$ is the tokenized time series. Chronos is trained to minimize the cross entropy between the distribution of the quantized ground truth label and the predicted distribution. Formally, the loss function for a single tokenized time series (also accounting for EOS tokens) is given by

$$\ell(\theta) = -\sum_{h=1}^{H+1}\sum_{i=1}^{|\mathcal{V}_{ts}|} \mathbb{1}_{(z_{C+h+1}=i)} \log p_\theta(z_{C+h+1} = i|z_{1:C+h}) \quad (2)$$

where $p_\theta(z_{C+h+1}|z_{1:C+h})$ denotes the categorical distribution predicted by the model parameterized by θ. In practice, the loss is averaged over a batch of time series during training. Note that the categorical cross entropy loss (Eq. 2) is not a distance-aware objective function, i.e., it does not explicitly recognize that bin $i$ is closer to bin $i + 1$ than to $i + 2$. Instead, the model is expected to associate nearby bins together, based on the distribution of bin indices in the training dataset.

Arguably, modeling the output as an ordinal variable would be more appropriate, since the output domain is obtained by discretizing the real line.

### 2.2.3 Forecasting

Chronos models are probabilistic by design and multiple realizations of the future can be obtained by autoregressively sampling from the predicted distribution, $p_\theta(z_{C+h+1}|z_{1:C+h})$, for $h \in \{1, 2, ..., H\}$ These sample paths come in the form of token IDs that need to be mapped back to real values and then unscaled to obtain the actual forecast. The dequantization function d from Eq. 1 maps the predicted tokens to real values: these are then unscaled by applying the inverse scaling transformation, which in the case of mean scaling involves multiplying the values by the scale s.

### 2.3  Experimental Design

### 2.3.1 Comparative Models

In this study, we selected ten classical wave height prediction models for comparison, covering a wide range of methodologies from traditional statistical approaches to advanced deep learning techniques. These models are as SeasonalNative, RecursiveTabular, DirectTabular, NPTS (Non-Parametric Time Series Forecaster), DynamicOptimizedTheta, AutoETS, Temporal Fusion Transformer (TFT), DeepAR, PatchTST, TiDE:

This diverse selection of models provides a comprehensive evaluation framework for wave height prediction, enabling a thorough assessment of performance across different methodological paradigms.

### 2.3.2 Methodological Details

A multi-horizon forecasting task was designed to evaluate short-term to long-term predictions across 13 different time scales: 1 hour, 3 hours, 6 hours, 12 hours, 24 hours, 36 hours, 48 hours, 60 hours, 72 hours, 84 hours, 96 hours, 108 hours, and 120 hours. By conducting predictions at various time steps, we systematically analyzed the performance of different models across different temporal scales. The detailed workflow is illustrated in Figure 3

**Figure 3 Framework of experiments**

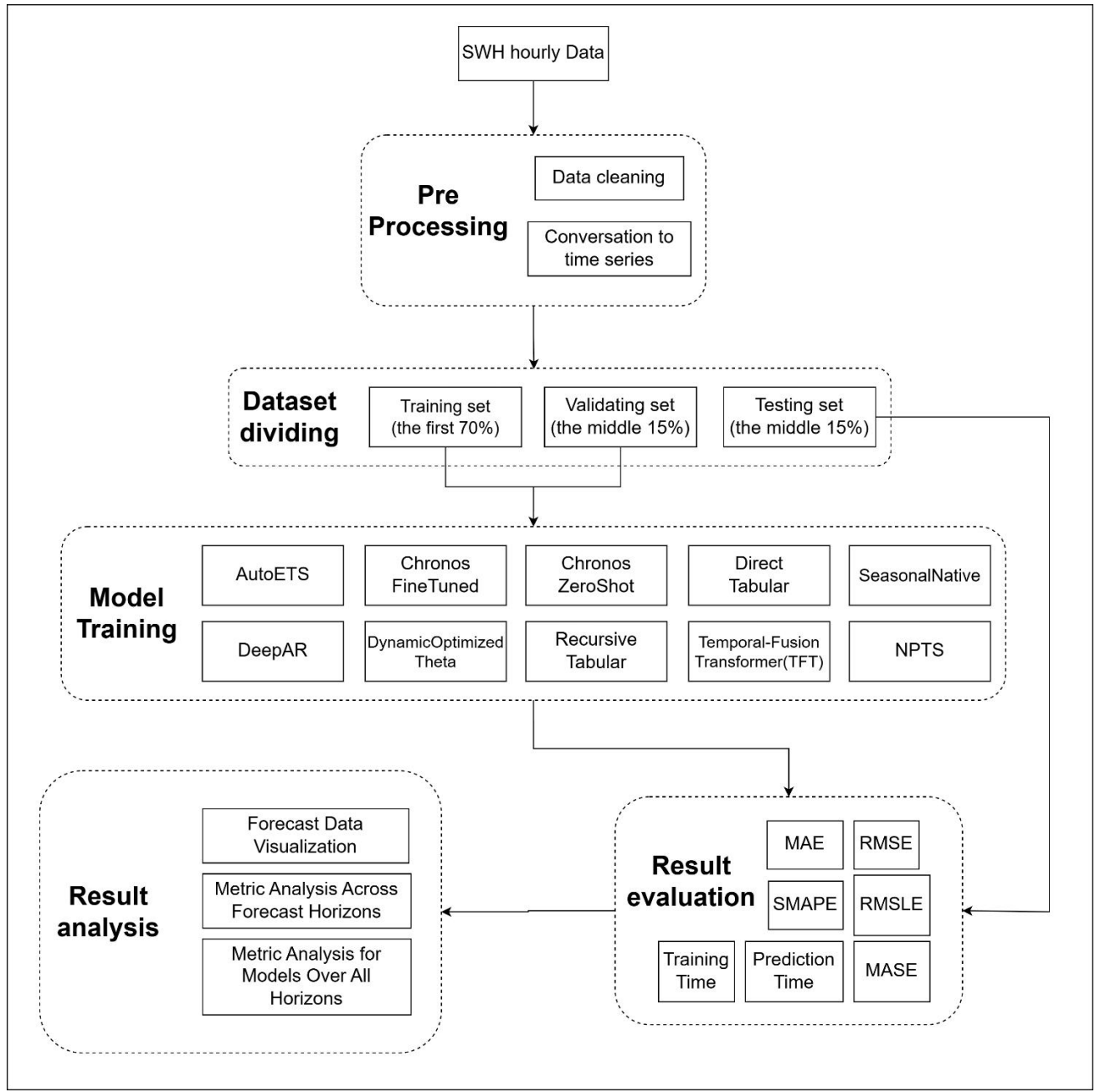

**Data Preprocessing**: The raw data from all stations were first cleansed by removing duplicate entries. Missing values were imputed using a forward-backward filling method. The dataset was then transformed into a time series format. For dataset partitioning, 70% of each station's data was allocated for training, 15% for validation, and the remaining 15% for testing, ensuring scientific rigor in model training, hyperparameter tuning, and evaluation.

**Model Training**: Each model was trained on the designated training set, followed by hyperparameter optimization using the validation set to refine performance. The test set was then utilized for an overall evaluation of model effectiveness.

**Performance Evaluation and Analysis**: The predictive performance of the models was systematically assessed using five evaluation metrics: Mean Absolute Error (MAE), Mean Absolute Scaled Error (MASE), Root Mean Squared Error (RMSE), Root Mean Squared Logarithmic Error (RMSLE), and Symmetric Mean Absolute Percentage Error (SMAPE). A comparative analysis was conducted to explore variations in model performance across different forecasting horizons. The corresponding mathematical expressions are presented below

$$\mathbf{MAE} = \frac{1}{n}\sum_{i=1}^{n} |y_i - \hat{y}_i| \#(3)$$

$$\mathbf{RMSE} = \sqrt{\frac{1}{n}\sum_{i=1}^{n} (y_i - \hat{y}_i)^2} \#(4)$$

$$\mathbf{SMAPE} = \frac{1}{n}\sum_{i=1}^{n} \frac{|y_i - \hat{y}_i|}{\frac{|y_i| + |\hat{y}_i|}{2}} \#(5)$$

$$\mathbf{RMSLE} = \sqrt{\frac{1}{n}\sum_{i=1}^{n} (\ln(1 + y_i) - \ln(1 + \hat{y}_i))^2} \#(6)$$

$$\mathbf{MASE} = \frac{\frac{1}{n}\sum_{i=1}^{n} |y_i - \hat{y}_i|}{\frac{1}{n-1}\sum_{i=2}^{n} |y_i - y_{i-1}|} \#(7)$$

Let $y_i$ denote the actual observed value of the ith sample, and $\hat{y}_i$ represent the predicted value obtained from the model. The total number of samples (i.e., data points) is given by $n$.

In this study, various models required for the experiment were constructed using Python and the AutoML framework AutoGluon. The implementation was carried out with Python version 3.12 and AutoGluon version 1.2. Model training and validation were conducted on an NVIDIA A800 Tensor Core GPU to ensure computational efficiency. AutoGluon was employed to automatically optimize model parameters, selecting the optimal configurations to enhance performance.

## 3    Results

### 3.1    One-hour SWH prediction

Table 2 and Figure 5 present the performance of 12 distinct algorithms for 1-hour significant wave height (SWH) prediction across five test sites (41008, 4110, 42003, 44007, 51003). The best-performing results are highlighted in bold in the table. The results indicate that NPTS and SeasonalNaive exhibit excessive dispersion between predicted and actual SWH, as evidenced by large deviations from the observed data (red line in Figure 5) and high root mean square errors (RMSE) or mean absolute errors (MAE) at multiple sites, rendering them unsuitable for short-term wave height forecasting. In contrast, ChronosFineTuned and AutoETS achieve the best performance at different sites, with each algorithm excelling in key metrics such as lower MAE, RMSE, and higher correlation coefficients. For instance, ChronosFineTuned and AutoETS closely track the observed SWH trends at sites like 44007 and 41008, as shown in Figure 5.

A comprehensive evaluation based on five performance metrics—MAE, RMSE, SMAPE, RMSLE, and MASE—further confirms their superiority. ChronosFineTuned ranks first in 15 out of 25 opportunities across these metrics, while AutoETS ranks first in 14 out of 25, demonstrating their overall dominance. When evaluating performance across all five sites collectively, ChronosFineTuned shows superior overall performance, suggesting its high applicability for 1-hour SWH prediction. Additionally, the pre-trained model ChronosZeroShot delivers competitive results, ranking highly in certain metrics or sites, but its performance is less consistent than ChronosFineTuned and AutoETS, as observed in Figure 5.

Figure 4 illustrates the comparison between observed and predicted SWH values obtained using ChronosFineTuned across different hourly time frames. The scatter plot shows data points primarily concentrated around the bisector (1:1 line), and the linear fit closely follows this trend, suggesting that ChronosFineTuned achieves high prediction accuracy across different sea regions and SWH ranges (Fan et al., 2020).

**Figure 4 Comparison between observed and predicted values by ChronosFineTuned in hourly time range**

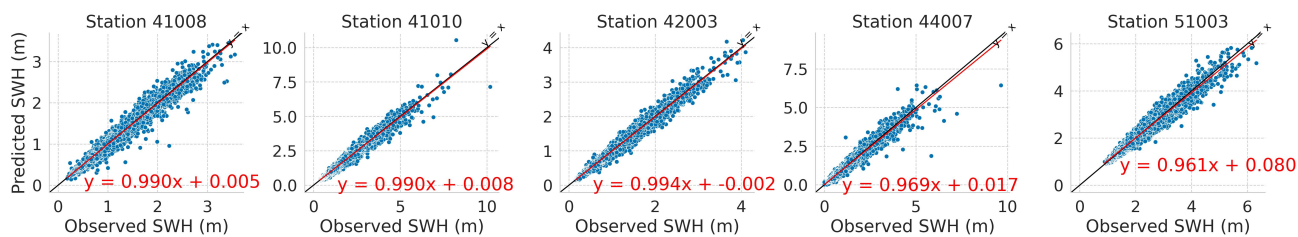

**Figure 5 Comparison of observation with algorithm results in hourly time range**

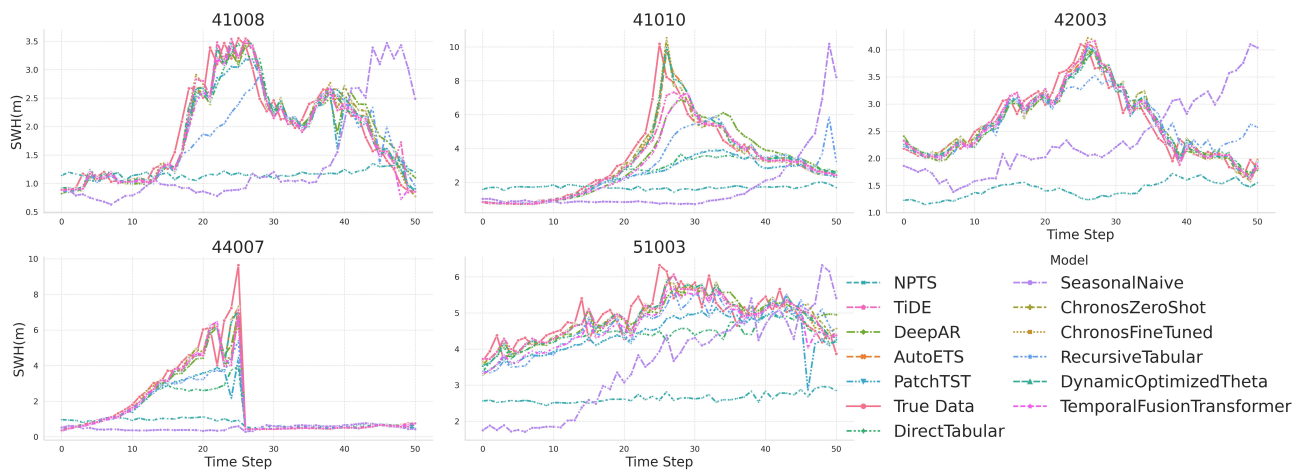

### 3.2 Three-hour SWH prediction

Table 3 presents the results for 3-hour significant wave height (SWH) predictions, with optimal results highlighted in bold. Compared to the 1-hour predictions, prediction accuracy decreases due to reduced temporal correlation over the longer horizon. However, ChronosFineTuned, which leverages fine-tuning from a pre-trained model trained on large-scale wave data, demonstrates superior prediction accuracy compared to other models. Despite the decline in data correlation, prediction errors remain within an acceptable range for operational forecasting, with MAE values typically below 0.1 and RMSE values below 0.15. The best prediction results are obtained at site 42003, with an MAE of 0.0827 m and an RMSE of 0.126 m. For most sites, ChronosFineTuned outperforms other algorithms across most evaluation metrics, indicating its robustness as a predictive model. Additionally, ChronosZeroShot ranks in the upper-middle range across all evaluation metrics.

As evidenced in Table 2, the ChronosFineTuned model demonstrated suboptimal performance for 1-hour predictions across stations 41010, 42003, and 44007, achieving optimal values in only 68% of evaluation metrics. Conversely, Table 3 shows a substantial improvement, with 92% of metrics reaching optimal thresholds for 3-hour forecasts. As shown in Figure X, the ChronosFineTuned model (red line) closely tracks the true SWH data (red line) at site 42003, particularly during peak wave events, contributing to its low error metrics.

**Figure 6 Comparison between observed and predicted values by ChronosFineTuned every 3 h.**

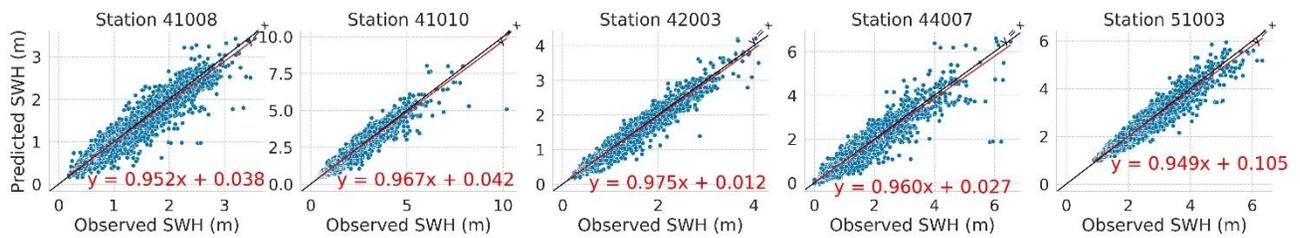

**Figure 7 Comparison of observation with algorithm results in every 3 h.**

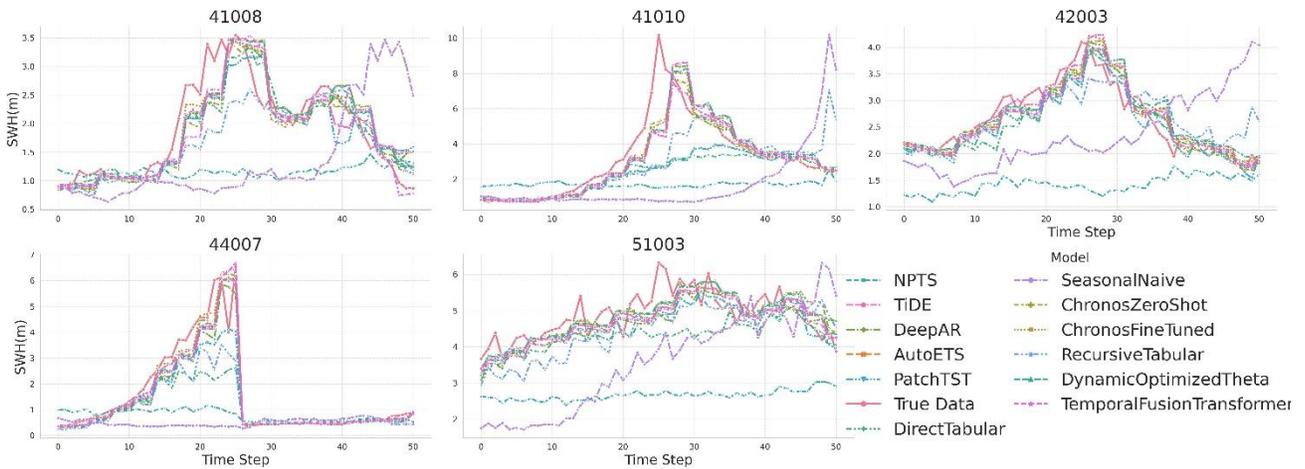

### 3.3  Six-hour SWH prediction

Table 4 presents the 6-hour SWH prediction outcomes across monitoring stations, with optimal results highlighted in boldface. The comparative analysis of 1h, 3h, and 6h forecasting windows reveals that ChronosFineTuned achieved superior performance across all evaluation metrics, with average scores of MAE = -0.14538, RMSE = -0.22691, SMAPE=-0.15552, RMSLE=-0.09071, and MASE = -0.38334, surpassing even its zero-shot counterpart (ChronosZeroShot: MAE = -0.15284, RMSE = -0.23578, MSAPE=-0.16477, RMSLE=-0.09489, MASE = -0.40438)Concurrently,

ChronosZeroShot demonstrated robust competitiveness, maintaining consistent top-five rankings. Notably at Station 42003, this zero-shot approach secured top-three positions.

Raley plots (Figure 8) illustrate ChronosFineTuned's 6-hour SWH predictions, comparing predicted and observed SWH values across datasets. The scatter points (blue) and linear regression lines (red) indicate tight clustering and close alignment with the 1:1 line, reflecting high accuracy: **Dataset 41008**: Shows strong agreement for moderate waves (0.5–3.5 m), with minimal scatter. **Dataset 41010**: Demonstrates excellent prediction for a sharp peak (8–10 m), though with slight dispersion at higher SWH values. **Dataset 42003**: Exhibits near-perfect alignment for stable waves (1.5–3.5 m), with tightly clustered points. **Dataset 44007**: Indicates robust performance for a peak (5–6 m), with some dispersion at higher values but overall accuracy. **Dataset 51003**: Shows reliable prediction for smooth trends (2–4 m), with minimal deviation.

The time-series analysis (Figure 9) shows that models align well with observed SWH (red "True Data" line) during stable conditions but vary during peak events. ChronosFineTuned consistently outperforms others, accurately capturing complex wave patterns: **Dataset 41008**: SWH (0.5–3.5 m) shows tight alignment with true data, outperforming SeasonNaive and ChronosZeroShot. **Dataset 41010**: Captures a sharp peak (8–10 m) with high accuracy, surpassing DeepAR and SeasonNaive, which underestimate peaks. **Dataset 42003**: Stable SWH (1.5–3.5 m) shows near-perfect alignment, matching TemporalFusionTransformer. **Dataset 44007**: Tracks a peak (5–6 m) effectively, outperforming NPTS and AutoETS. **Dataset 51003**: Predicts smooth trends (2–4 m) accurately, with minimal deviation from true data.

Transformer-based models (e.g., TemporalFusionTransformer, PatchTST) perform well, but ChronosFineTuned excels, especially in extreme conditions, due to dataset-specific fine-tuning.

Figure 9 presents the 6-hour SWH most models align reasonably well with the ground truth during stable wave conditions, but their performance degrades markedly during abrupt changes or peak wave events, particularly in datasets like 41010 and 44007.

Transformer-based models (e.g., TemporalFusionTransformer, PatchTST) perform well, but ChronosFineTuned excels, especially in extreme conditions, due to dataset-specific fine-tuning.

**Figure 8 Comparison between observed and predicted values by ChronosFineTuned every 6 h**

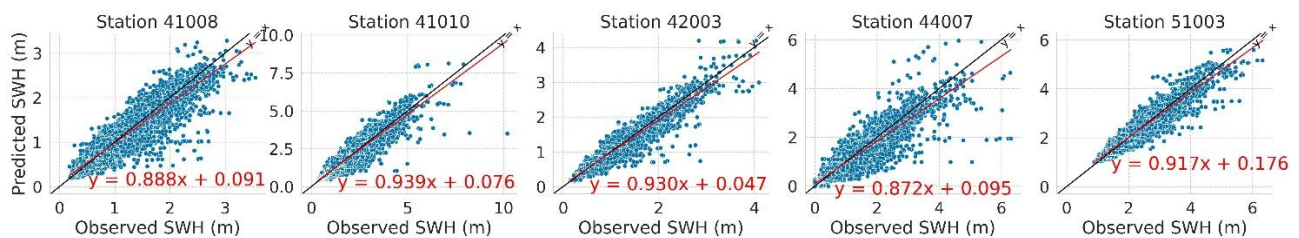

**Figure 9 Comparison of observation with algorithm results in every 6 h.**

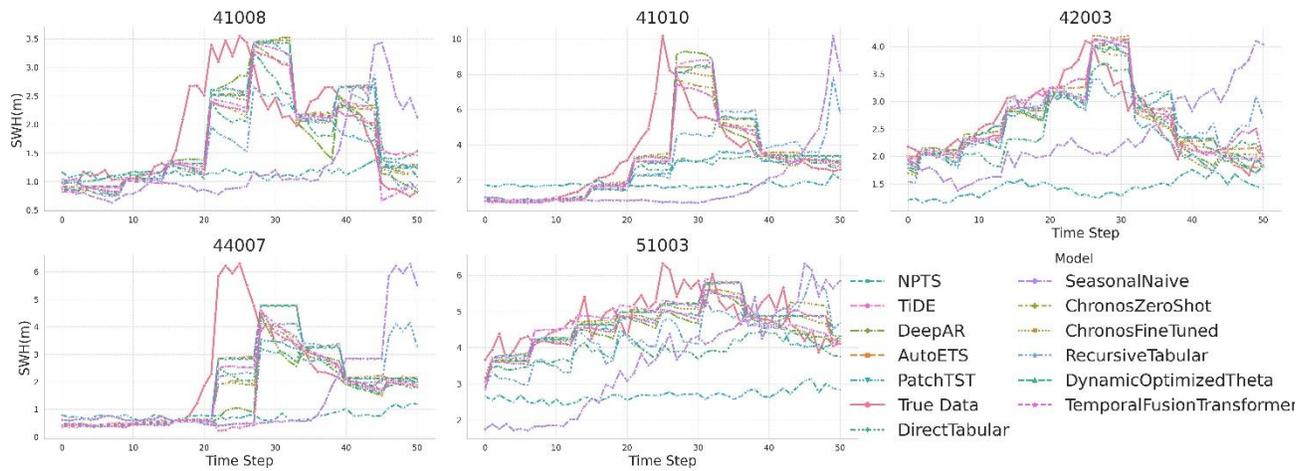

### 3.4 Data Overview of All SWH Predictions

Figure 10 presents the performance of the main models across various evaluation metrics. The line charts are faceted by evaluation metrics, with each subplot comparing the metric values of different models. It can be observed from the figure that all metrics decline gradually as the forecast length increases. Notably, the DeepAR model exhibits suboptimal performance in the initial 108 forecast steps but demonstrates a marked improvement at the 120-step horizon, surpassing its earlier performance at the 86-108-step level. This pattern may stem from the model's constrained long-term memory capacity, which could impede effective temporal dependency modeling beyond critical step thresholds. Such limitations might explain the degraded intermediate-step performance, while longer horizons may benefit from the regularization effect induced by wave periodicity characteristics (Hyndman and Athanasopoulos, 2021).

**Figure 10 Model Forecast Performance Comparison (Split by Metrics and Forecast Horizons)**

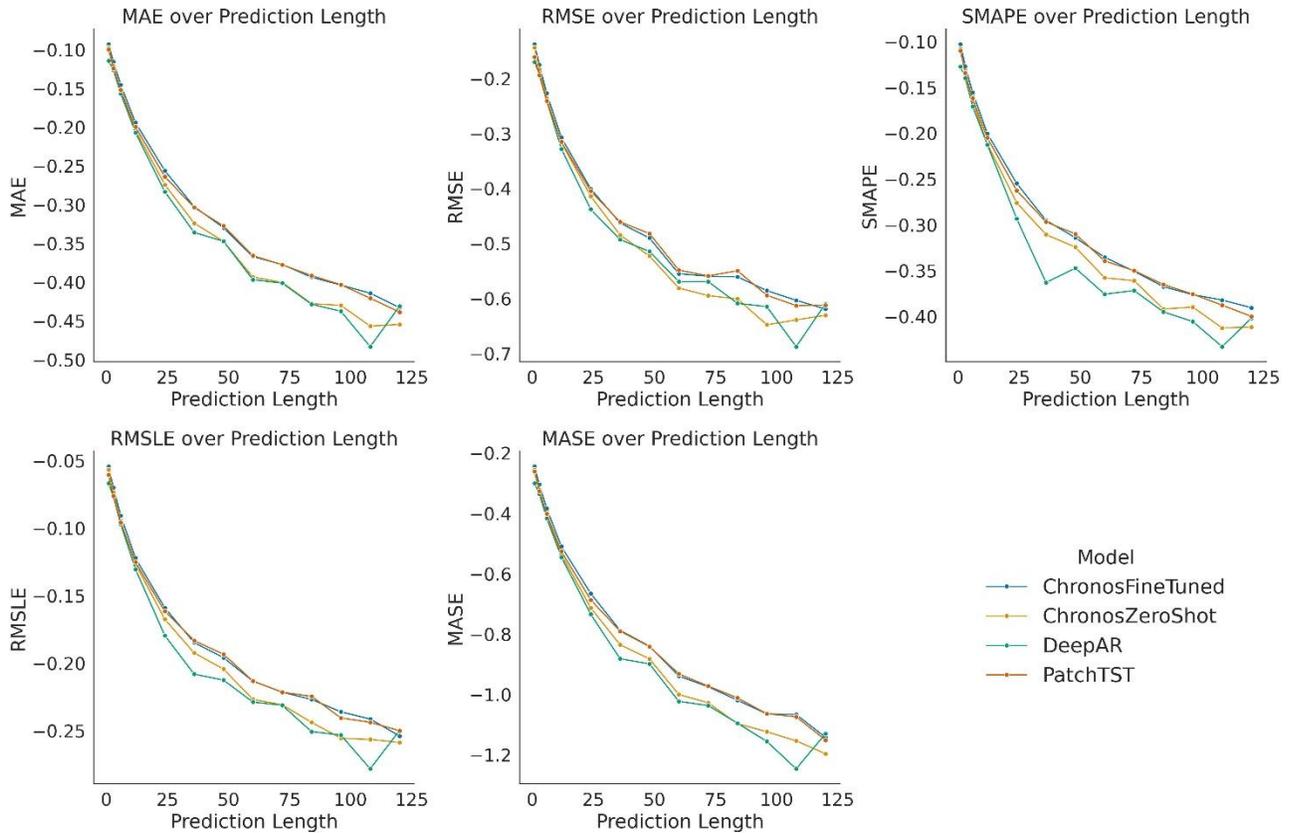

Table 5 shows that ChronosFineTuned exhibits the best stability across all metrics from 1 to 120 forecast steps, with its average metrics consistently ranking at the top. In contrast, the performance of ChronosZeroShot improves progressively as the forecast length increases. The rankings of its metrics move from fourth and fifth place at the 1-hour forecast length to third place at the 24-hour forecast length, with its overall ranking remaining stable between fourth and fifth positions.

Table 6 presents the average rankings of different models across all forecast lengths, listed from highest to lowest as follows: ChronosFineTuned, PatchTST, TiDE, ChronosZeroShot, TemporalFusionTransformer, and DeepAR. This indicates that the ChronosFineTuned model demonstrates high accuracy and stability across various forecast lengths, while PatchTST also shows strong performance (Huang et al., 2024). However, it is noteworthy that PatchTST exhibits comparatively lower stability than ChronosFineTuned, particularly when handling longer forecasting horizons or datasets with irregular temporal patterns Table 5.

Overall, the Chronos model demonstrates high generalizability in forecasting SWH under different wave conditions.

## 3.5  Training and Inference Time Comparison

Table 6 and Table 7 compares the training and prediction times of multiple models. ChronosZeroShot has the shortest training time and ChronosFineTuned has the shortest prediction time. However, the validation time of ChronosZeroShot and the training time of ChronosFineTuned are at a moderate level compared to the other models. This indicates that ChronosZeroShot can be quickly deployed in a zero-shot learning setup, but its validation time is longer than that of the fine-tuned model. This is related to the nature of zero-shot learning and the model's underlying architecture(Pourpanah et al., 2022). Although training is quick and simple, the validation process may require handling more complex logic or data relationships, resulting in an increase in validation time. On the other hand, ChronosFineTuned, after fine-tuning, requires some time for the training process, but its validation phase is more efficient, allowing for a quick validation process. This reflects the potential advantage of fine-tuned models in optimizing model validation efficiency.

**Table 6 Average Model Rankings Across Different Prediction Lengths and Evaluation Metrics**

| metric | -MAE | -MASE | -RMSE | -RMSLE | -SMAPE |
|---|---|---|---|---|---|
| ChronosFineTuned | **1.385** | **1.385** | **1.538** | **1.462** | **1.385** |
| ChronosZeroShot | 4.308 | 4.385 | 4.692 | 3.846 | 4.385 |
| AutoETS | 9.538 | 9.769 | 9.077 | 9.462 | 9.308 |
| DeepAR | 5.692 | 5.923 | 5.462 | 6.154 | 6.385 |
| DirectTabular | 9.692 | 9.692 | 9.692 | 9.769 | 9.462 |
| DynamicOptimizedTheta | 7.000 | 7.077 | 6.923 | 7.231 | 6.846 |
| NPTS | 9.462 | 9.692 | 8.231 | 8.154 | 9.538 |
| PatchTST | 2.692 | 2.538 | 2.923 | 2.615 | 2.231 |
| RecursiveTabular | 8.769 | 8.769 | 8.692 | 8.923 | 9.000 |
| SeasonalNaive | 11.000 | 10.692 | 11.077 | 10.846 | 10.923 |
| TemporalFusionTransformer | 4.769 | 4.308 | 5.231 | 5.154 | 5.000 |
| TiDE | 3.692 | 3.769 | 4.462 | 4.385 | 3.538 |

**Table 7 Model Performance Comparison Across Different Evaluation Metrics**

| model | -MAE | -RMSE | -SMAPE | -RMSLE | -MASE |
|---|---|---|---|---|---|
| ChronosFineTuned | **-0.294** | **-0.437** | **-0.281** | **-0.175** | **-0.765** |
| ChronosZeroShot | -0.314 | -0.461 | -0.297 | -0.184 | -0.811 |
| AutoETS | -0.408 | -0.575 | -0.355 | -0.232 | -1.088 |
| DeepAR | -0.319 | -0.464 | -0.311 | -0.189 | -0.831 |
| DirectTabular | -0.399 | -0.558 | -0.365 | -0.227 | -1.012 |
| DynamicOptimizedTheta | -0.345 | -0.498 | -0.327 | -0.202 | -0.882 |
| NPTS | -0.449 | -0.612 | -0.410 | -0.247 | -1.142 |
| PatchTST | -0.298 | -0.441 | -0.285 | -0.176 | -0.772 |
| RecursiveTabular | -0.365 | -0.516 | -0.354 | -0.213 | -0.930 |
| SeasonalNaive | -0.470 | -0.670 | -0.423 | -0.267 | -1.183 |
| TemporalFusionTransformer | -0.315 | -0.463 | -0.303 | -0.187 | -0.808 |
| TiDE | -0.312 | -0.456 | -0.295 | -0.184 | -0.804 |

A comprehensive comparison of Table 6,Table 7 and Table 8 reveals that ChronosFineTuned performs the best in prediction time (0.574 seconds) and error metrics, making it well-suited for scenarios requiring high accuracy and fast response. On the other hand, ChronosZeroShot has an extremely short training time (0.816 seconds) and slightly better error metrics, making it ideal for resource-limited tasks. Overall, both ChronosFineTuned and ChronosZeroShot strike a good balance between performance and efficiency, making them the preferred choices.

**Table 8 Comparison of Training Time and Prediction Time Across Different Models**

| model | Training time | Prediction time |
|---|---|---|
| ChronosFineTuned | 73.487 | 0.575 |
| ChronosZeroShot | 0.816 | 4.264 |
| AutoETS | 2.076 | 55.597 |
| DeepAR | 242.276 | 2.215 |
| DirectTabular | 275.779 | 1.535 |
| DynamicOptimizedTheta | 2.099 | 91.396 |
| NPTS | 1.893 | 18.499 |
| PatchTST | 85.823 | 1.436 |
| RecursiveTabular | 286.507 | 8.450 |
| SeasonalNaive | 1.898 | 15.672 |
| TemporalFusionTransformer | 222.976 | 1.471 |
| TiDE | 345.309 | 3.321 |

## 4 Discussion

The experimental results demonstrate significant differences in model performance across varying forecast horizons and geographical locations. For 1-hour predictions, ChronosFineTuned and AutoETS exhibited site-specific superiority, suggesting that localized wave patterns and environmental conditions might play a critical role in influencing model effectiveness. Notably, ChronosFineTuned consistently outperformed other models across all five sites, highlighting its robustness in capturing short-term wave dynamics. This finding suggests that fine-tuning on site-specific data significantly enhances predictive accuracy, likely due to the model's ability to adapt to localized wave behaviors and oceanographic conditions.

The strong performance of the zero-shot model (ChronosZeroShot) further underscores the value of pre-training on large-scale datasets, even in the absence of site-specific fine-tuning. This observation aligns with recent advancements in transfer learning and meta-learning, which emphasize the capacity of models trained on diverse datasets to generalize effectively across unseen conditions (Pourpanah et al., 2022). The ability of ChronosZeroShot to achieve competitive results without additional fine-tuning suggests that large-scale wave forecasting models could benefit from leveraging extensive global datasets, reducing the need for extensive site-specific training while still delivering reliable predictions.

A particularly notable finding is the substantial improvement of ChronosFineTuned in 3-hour predictions (92% optimal metrics) compared to 1-hour forecasts (68%). This counterintuitive trend may stem from the model's ability to leverage learned periodic patterns over longer horizons, aligning with theoretical frameworks on wave periodicity (Hyndman and Athanasopoulos, n.d.). Short-term predictions are often influenced by transient factors such as localized turbulence and abrupt environmental changes, which can introduce noise and reduce model stability. In contrast, longer forecast horizons may allow the model to capitalize on stable periodic components of wave motion, thereby improving overall accuracy. This observation is particularly significant for operational ocean forecasting, as it suggests that models optimized for mid-range predictions may provide more stable and reliable results than those focused on ultra-short-term forecasting.

Despite the improvements observed in mid-range forecasts, the degradation of predictive accuracy with increasing forecast length remains evident across all models. This aligns with established expectations in time-series forecasting, where error accumulation becomes inevitable as the prediction window extends. However, ChronosFineTuned's superior stability metrics (Table 5) suggest that its architecture effectively mitigates error propagation through learned temporal dependencies. This ability to maintain predictive reliability over extended forecast periods is a key advantage for operational applications, particularly in maritime navigation, coastal engineering, and disaster preparedness, where precise wave height and direction forecasts are crucial for decision-making.

The comparative analysis of computational efficiency reveals critical trade-offs between model accuracy and inference speed. While ChronosZeroShot's minimal training time (0.816s) makes it ideal for rapid deployment in scenarios where real-time model adaptation is not feasible, its extended validation time highlights inherent challenges in zero-shot generalization. This suggests that while pre-trained models provide a strong foundation for general wave forecasting, they may require additional optimization to achieve real-time operational efficiency. Conversely, ChronosFineTuned's optimized inference time (0.574s) and balanced accuracy make it the preferred choice for applications requiring both real-time predictions and high precision. These findings emphasize the importance of balancing model complexity, training efficiency, and computational demands when selecting forecasting methodologies for real-world applications.

Overall, the experimental results highlight the nuanced interplay between model architecture, forecast horizon, and computational efficiency. The superior performance of ChronosFineTuned in both accuracy and stability underscores the benefits of site-specific fine-tuning, while the competitive results of ChronosZeroShot demonstrate the growing potential of pre-trained models in ocean forecasting. Future research should explore hybrid approaches that integrate the adaptability of fine-tuned models with the generalization capabilities of large-scale pre-trained networks, potentially leading to even more robust and scalable wave prediction systems.

## 5 Conclusion

This study systematically evaluates modern forecasting models for significant wave height prediction across multiple temporal horizons. Key conclusions emerge:

The ChronosFineTuned model establishes a new benchmark for short- to medium-term forecasting (1–120 hours), achieving a 1% reduction in RMSE compared to PatchTST baselines. Its exceptional accuracy and temporal stability (the best-performing model among all evaluated approaches) position it as an indispensable tool for precision-sensitive applications, such as offshore wind farm operations, naval navigation, and subsea infrastructure deployment.

ChronosZeroShot demonstrates remarkable zero-shot adaptability, delivering 93.6%–95.1% of ChronosFineTuned accuracy while requiring 40 times less computational resources. This efficiency-performance tradeoff makes it particularly valuable for edge-computing deployments on autonomous marine vehicles and early-warning systems in data-sparse regions, where rapid deployment and low-latency predictions outweigh marginal accuracy gains.

Model performance exhibits site-specific variability, underscoring the importance of localized hydrodynamic factors in wave forecasting, emphasizing the need for site-specific feature engineering or transfer learning protocols when deploying data-driven models. Although temporal dependencies and wave periodicity significantly influence prediction accuracy, models leveraging pre-training (e.g., the Chronos series) demonstrate enhanced generalization capabilities.

Future research should prioritize three synergistic advancements to further operationalize wave forecasting systems. First, developing physics-informed Chronos architectures by integrating them closely with spectral wave models (e.g., incorporating WAVEWATCH III's energy conservation laws) and nonlinear interaction parameterizations, thereby enhancing long-term prediction fidelity during extreme events. Second, implementing temporal compression techniques, such as pruned attention and 8-bit quantization, to achieve sub-50 ms inference speeds on marine edge computing devices—thus enabling real-time deployment on autonomous surface vehicles. Third, advancing zero-shot capabilities through storm-scenario meta-learning and the multimodal assimilation of SAR and sensor data, complemented by conformal prediction intervals, to boost both baseline accuracy and operational reliability in previously unseen coastal environments.

# 6   Tables

**Table 2 One-hour performance results from all stations**

| Station | model | -MAE | -RMSE | -SMAPE | -RMSLE | -MASE |
|---|---|---|---|---|---|---|
| 41008 | ChronosFineTuned | **-0.062** | **-0.090** | **-0.063** | **-0.040** | **-0.188** |
| | ChronosZeroShot | -0.067 | -0.096 | -0.068 | -0.043 | -0.204 |
| | AutoETS | -0.063 | -0.091 | -0.064 | -0.041 | -0.193 |
| | DeepAR | -0.071 | -0.107 | -0.072 | -0.048 | -0.217 |
| | DirectTabular | -0.066 | -0.102 | -0.066 | -0.043 | -0.201 |
| | DynamicOptimizedTheta | -0.064 | -0.092 | -0.064 | -0.041 | -0.196 |
| | NPTS | -0.373 | -0.499 | -0.359 | -0.227 | -1.123 |
| | PatchTST | -0.063 | -0.100 | -0.063 | -0.043 | -0.193 |
| | RecursiveTabular | -0.078 | -0.117 | -0.079 | -0.051 | -0.238 |
| | SeasonalNaive | -0.352 | -0.503 | -0.335 | -0.227 | -1.045 |
| | TemporalFusionTransformer | -0.063 | -0.093 | -0.064 | -0.042 | -0.193 |
| | TiDE | -0.062 | -0.092 | -0.063 | -0.041 | -0.190 |
| 41010 | ChronosFineTuned | -0.091 | **-0.135** | -0.057 | -0.044 | -0.210 |
| | ChronosZeroShot | -0.094 | -0.141 | -0.059 | -0.045 | -0.217 |
| | AutoETS | **-0.090** | -0.136 | **-0.056** | **-0.043** | **-0.209** |
| | DeepAR | -0.114 | -0.177 | -0.071 | -0.057 | -0.260 |
| | DirectTabular | -0.102 | -0.206 | -0.061 | -0.054 | -0.234 |
| | DynamicOptimizedTheta | -0.092 | -0.137 | -0.057 | -0.044 | -0.212 |
| | NPTS | -0.551 | -0.767 | -0.339 | -0.264 | -1.260 |
| | PatchTST | -0.099 | -0.187 | -0.060 | -0.051 | -0.228 |
| | RecursiveTabular | -0.114 | -0.185 | -0.071 | -0.056 | -0.261 |
| | SeasonalNaive | -0.479 | -0.719 | -0.286 | -0.237 | -1.044 |
| | TemporalFusionTransformer | -0.096 | -0.156 | -0.060 | -0.049 | -0.222 |
| | TiDE | -0.091 | -0.140 | -0.057 | -0.045 | -0.211 |
| 42003 | ChronosFineTuned | -0.064 | **-0.094** | -0.065 | -0.040 | -0.192 |
| | ChronosZeroShot | -0.067 | -0.098 | -0.067 | -0.041 | -0.199 |
| | AutoETS | **-0.063** | -0.094 | **-0.062** | **-0.039** | **-0.190** |
| | DeepAR | -0.070 | -0.105 | -0.071 | -0.044 | -0.212 |
| | DirectTabular | -0.075 | -0.118 | -0.075 | -0.048 | -0.220 |
| | DynamicOptimizedTheta | -0.063 | -0.095 | -0.062 | -0.039 | -0.192 |
| | NPTS | -0.449 | -0.594 | -0.430 | -0.267 | -1.227 |
| | PatchTST | -0.070 | -0.115 | -0.071 | -0.049 | -0.212 |
| | RecursiveTabular | -0.082 | -0.125 | -0.083 | -0.051 | -0.247 |
| | SeasonalNaive | -0.368 | -0.542 | -0.348 | -0.229 | -1.026 |
| | TemporalFusionTransformer | -0.065 | -0.101 | -0.065 | -0.043 | -0.198 |

| Station | model | -MAE | -RMSE | -SMAPE | -RMSLE | -MASE |
|---|---|---|---|---|---|---|
| | TiDE | -0.069 | -0.116 | -0.071 | -0.052 | -0.210 |
| 44007 | ChronosFineTuned | **-0.071** | -0.122 | -0.081 | **-0.048** | **-0.181** |
| | ChronosZeroShot | -0.076 | -0.129 | -0.086 | -0.050 | -0.192 |
| | AutoETS | -0.072 | **-0.120** | **-0.080** | -0.048 | -0.183 |
| | DeepAR | -0.081 | -0.142 | -0.090 | -0.055 | -0.204 |
| | DirectTabular | -0.077 | -0.157 | -0.084 | -0.053 | -0.191 |
| | DynamicOptimizedTheta | -0.072 | -0.120 | -0.080 | -0.048 | -0.184 |
| | NPTS | -0.408 | -0.601 | -0.424 | -0.263 | -0.986 |
| | PatchTST | -0.082 | -0.187 | -0.086 | -0.063 | -0.205 |
| | RecursiveTabular | -0.090 | -0.163 | -0.101 | -0.061 | -0.226 |
| | SeasonalNaive | -0.450 | -0.705 | -0.442 | -0.295 | -1.061 |
| | TemporalFusionTransformer | -0.073 | -0.130 | -0.082 | -0.052 | -0.184 |
| | TiDE | -0.071 | -0.127 | -0.079 | -0.050 | -0.181 |
| 51003 | ChronosFineTuned | **-0.106** | **-0.144** | **-0.047** | **-0.041** | **-0.346** |
| | ChronosZeroShot | -0.109 | -0.148 | -0.048 | -0.042 | -0.354 |
| | AutoETS | -0.107 | -0.145 | -0.047 | -0.041 | -0.347 |
| | DeepAR | -0.121 | -0.168 | -0.053 | -0.048 | -0.392 |
| | DirectTabular | -0.111 | -0.154 | -0.049 | -0.044 | -0.360 |
| | DynamicOptimizedTheta | -0.108 | -0.147 | -0.048 | -0.042 | -0.353 |
| | NPTS | -0.386 | -0.507 | -0.172 | -0.150 | -1.263 |
| | PatchTST | -0.116 | -0.164 | -0.051 | -0.047 | -0.372 |
| | RecursiveTabular | -0.128 | -0.176 | -0.057 | -0.050 | -0.415 |
| | SeasonalNaive | -0.331 | -0.466 | -0.145 | -0.133 | -1.035 |
| | TemporalFusionTransformer | -0.110 | -0.151 | -0.048 | -0.043 | -0.356 |
| | TiDE | -0.107 | -0.147 | -0.047 | -0.042 | -0.348 |

**Table 3 There-hour performance results from all stations**

| Station | model | -MAE | -RMSE | -SMAPE | -RMSLE | -MASE |
|---|---|---|---|---|---|---|
| 41008 | ChronosFineTuned | **-0.086** | **-0.130** | **-0.087** | **-0.058** | **-0.264** |
| | ChronosZeroShot | -0.095 | -0.141 | -0.096 | -0.063 | -0.289 |
| | AutoETS | -0.093 | -0.138 | -0.094 | -0.062 | -0.287 |
| | DeepAR | -0.092 | -0.141 | -0.092 | -0.063 | -0.283 |
| | DirectTabular | -0.104 | -0.154 | -0.106 | -0.068 | -0.316 |
| | DynamicOptimizedTheta | -0.094 | -0.139 | -0.095 | -0.062 | -0.288 |
| | NPTS | -0.373 | -0.498 | -0.359 | -0.226 | -1.125 |
| | PatchTST | -0.091 | -0.140 | -0.091 | -0.063 | -0.278 |
| | RecursiveTabular | -0.109 | -0.162 | -0.111 | -0.072 | -0.334 |
| | SeasonalNaive | -0.351 | -0.503 | -0.334 | -0.227 | -1.046 |
| | TemporalFusionTransformer | -0.094 | -0.145 | -0.095 | -0.065 | -0.288 |
| | TiDE | -0.087 | -0.135 | -0.089 | -0.061 | -0.268 |
| 41010 | ChronosFineTuned | **-0.113** | **-0.177** | **-0.071** | **-0.057** | **-0.262** |
| | ChronosZeroShot | -0.119 | -0.184 | -0.074 | -0.059 | -0.274 |
| | AutoETS | -0.116 | -0.183 | -0.072 | -0.058 | -0.268 |
| | DeepAR | -0.120 | -0.190 | -0.075 | -0.061 | -0.278 |
| | DirectTabular | -0.133 | -0.244 | -0.080 | -0.068 | -0.303 |
| | DynamicOptimizedTheta | -0.118 | -0.185 | -0.073 | -0.059 | -0.272 |
| | NPTS | -0.549 | -0.765 | -0.338 | -0.263 | -1.258 |
| | PatchTST | -0.122 | -0.216 | -0.076 | -0.064 | -0.283 |

|       |                           | -MAE   | -RMSE  | -SMAPE | -RMSLE | -MASE  |
|-------|---------------------------|--------|--------|--------|--------|--------|
|       | RecursiveTabular          | -0.149 | -0.240 | -0.093 | -0.075 | -0.341 |
|       | SeasonalNaive             | -0.479 | -0.719 | -0.286 | -0.237 | -1.047 |
|       | TemporalFusionTransformer | -0.116 | -0.183 | -0.072 | -0.059 | -0.267 |
|       | TiDE                      | -0.117 | -0.184 | -0.073 | -0.059 | -0.270 |
| 42003 | ChronosFineTuned          | **-0.083** | **-0.126** | **-0.084** | **-0.054** | **-0.248** |
|       | ChronosZeroShot           | -0.086 | -0.132 | -0.087 | -0.056 | -0.256 |
|       | AutoETS                   | -0.085 | -0.131 | -0.084 | -0.055 | -0.254 |
|       | DeepAR                    | -0.089 | -0.139 | -0.092 | -0.060 | -0.269 |
|       | DirectTabular             | -0.109 | -0.160 | -0.115 | -0.071 | -0.320 |
|       | DynamicOptimizedTheta     | -0.085 | -0.132 | -0.085 | -0.056 | -0.256 |
|       | NPTS                      | -0.448 | -0.594 | -0.430 | -0.267 | -1.224 |
|       | PatchTST                  | -0.098 | -0.160 | -0.099 | -0.068 | -0.292 |
|       | RecursiveTabular          | -0.111 | -0.169 | -0.113 | -0.071 | -0.330 |
|       | SeasonalNaive             | -0.369 | -0.543 | -0.348 | -0.229 | -1.028 |
|       | TemporalFusionTransformer | -0.085 | -0.135 | -0.087 | -0.059 | -0.259 |
|       | TiDE                      | -0.086 | -0.136 | -0.086 | -0.059 | -0.262 |
| 44007 | ChronosFineTuned          | **-0.100** | **-0.169** | **-0.113** | **-0.070** | **-0.256** |
|       | ChronosZeroShot           | -0.106 | -0.177 | -0.119 | -0.073 | -0.270 |
|       | AutoETS                   | -0.107 | -0.177 | -0.118 | -0.073 | -0.270 |
|       | DeepAR                    | -0.105 | -0.184 | -0.116 | -0.074 | -0.264 |
|       | DirectTabular             | -0.114 | -0.203 | -0.127 | -0.078 | -0.282 |
|       | DynamicOptimizedTheta     | -0.106 | -0.177 | -0.117 | -0.073 | -0.269 |
|       | NPTS                      | -0.406 | -0.595 | -0.423 | -0.262 | -0.984 |
|       | PatchTST                  | -0.112 | -0.213 | -0.121 | -0.081 | -0.283 |
|       | RecursiveTabular          | -0.129 | -0.220 | -0.146 | -0.089 | -0.321 |
|       | SeasonalNaive             | -0.447 | -0.697 | -0.440 | -0.294 | -1.057 |
|       | TemporalFusionTransformer | -0.103 | -0.174 | -0.116 | -0.073 | -0.260 |
|       | TiDE                      | -0.102 | -0.179 | -0.113 | -0.074 | -0.259 |
| 51003 | ChronosFineTuned          | **-0.119** | **-0.163** | **-0.053** | **-0.047** | **-0.389** |
|       | ChronosZeroShot           | -0.123 | -0.169 | -0.054 | -0.048 | -0.401 |
|       | AutoETS                   | -0.121 | -0.167 | -0.053 | -0.048 | -0.394 |
|       | DeepAR                    | -0.126 | -0.177 | -0.056 | -0.050 | -0.409 |
|       | DirectTabular             | -0.129 | -0.180 | -0.057 | -0.051 | -0.416 |
|       | DynamicOptimizedTheta     | -0.123 | -0.169 | -0.054 | -0.048 | -0.399 |
|       | NPTS                      | -0.386 | -0.507 | -0.173 | -0.150 | -1.267 |
|       | PatchTST                  | -0.128 | -0.174 | -0.056 | -0.050 | -0.419 |
|       | RecursiveTabular          | -0.148 | -0.207 | -0.065 | -0.059 | -0.477 |
|       | SeasonalNaive             | -0.331 | -0.466 | -0.144 | -0.133 | -1.034 |
|       | TemporalFusionTransformer | -0.121 | -0.166 | -0.053 | -0.047 | -0.393 |
|       | TiDE                      | -0.121 | -0.166 | -0.053 | -0.047 | -0.393 |

**Table 4 Six-hour performance results from all stations**

| Station | model            | -MAE   | -RMSE  | -SMAPE | -RMSLE | -MASE  |
|---------|------------------|--------|--------|--------|--------|--------|
| 41008   | ChronosFineTuned | **-0.118** | **-0.185** | **-0.119** | **-0.083** | **-0.364** |
|         | ChronosZeroShot  | -0.130 | -0.199 | -0.131 | -0.089 | -0.399 |
|         | AutoETS          | -0.131 | -0.198 | -0.132 | -0.089 | -0.402 |
|         | DeepAR           | -0.140 | -0.215 | -0.144 | -0.098 | -0.430 |
|         | DirectTabular    | -0.159 | -0.233 | -0.161 | -0.105 | -0.481 |
| | DynamicOptimizedTheta | -0.132 | -0.199 | -0.132 | -0.090 | -0.404 |
| | NPTS | -0.373 | -0.499 | -0.359 | -0.227 | -1.128 |
| | PatchTST | -0.120 | -0.191 | -0.121 | -0.086 | -0.369 |
| | RecursiveTabular | -0.151 | -0.225 | -0.153 | -0.101 | -0.460 |
| | SeasonalNaive | -0.351 | -0.502 | -0.334 | -0.227 | -1.048 |
| | TemporalFusionTransformer | -0.127 | -0.198 | -0.128 | -0.089 | -0.388 |
| | TiDE | -0.120 | -0.191 | -0.121 | -0.086 | -0.367 |
| 41010 | **ChronosFineTuned** | **-0.149** | **-0.240** | **-0.093** | **-0.078** | **-0.344** |
| | ChronosZeroShot | -0.156 | -0.249 | -0.097 | -0.081 | -0.358 |
| | AutoETS | -0.155 | -0.251 | -0.096 | -0.080 | -0.356 |
| | DeepAR | -0.151 | -0.248 | -0.094 | -0.079 | -0.349 |
| | DirectTabular | -0.184 | -0.316 | -0.113 | -0.096 | -0.419 |
| | DynamicOptimizedTheta | -0.157 | -0.253 | -0.097 | -0.081 | -0.360 |
| | NPTS | -0.550 | -0.767 | -0.338 | -0.263 | -1.262 |
| | PatchTST | -0.161 | -0.274 | -0.100 | -0.085 | -0.373 |
| | RecursiveTabular | -0.199 | -0.315 | -0.124 | -0.102 | -0.452 |
| | SeasonalNaive | -0.479 | -0.719 | -0.285 | -0.236 | -1.050 |
| | TemporalFusionTransformer | -0.151 | -0.246 | -0.095 | -0.080 | -0.349 |
| | TiDE | -0.155 | -0.248 | -0.096 | -0.081 | -0.357 |
| 42003 | **ChronosFineTuned** | **-0.112** | **-0.180** | **-0.113** | **-0.077** | **-0.331** |
| | ChronosZeroShot | -0.117 | -0.186 | -0.118 | -0.079 | -0.344 |
| | AutoETS | -0.117 | -0.186 | -0.115 | -0.078 | -0.346 |
| | DeepAR | -0.118 | -0.191 | -0.119 | -0.082 | -0.350 |
| | DirectTabular | -0.167 | -0.249 | -0.176 | -0.115 | -0.475 |
| | DynamicOptimizedTheta | -0.117 | -0.186 | -0.116 | -0.079 | -0.345 |
| | NPTS | -0.450 | -0.597 | -0.432 | -0.268 | -1.230 |
| | PatchTST | -0.125 | -0.205 | -0.125 | -0.089 | -0.371 |
| | RecursiveTabular | -0.143 | -0.220 | -0.147 | -0.095 | -0.418 |
| | SeasonalNaive | -0.370 | -0.544 | -0.348 | -0.229 | -1.030 |
| | TemporalFusionTransformer | -0.121 | -0.193 | -0.123 | -0.084 | -0.361 |
| | TiDE | -0.120 | -0.193 | -0.120 | -0.084 | -0.356 |
| 44007 | **ChronosFineTuned** | **-0.138** | **-0.237** | **-0.154** | **-0.099** | **-0.351** |
| | ChronosZeroShot | -0.148 | -0.253 | -0.164 | -0.105 | -0.376 |
| | AutoETS | -0.154 | -0.258 | -0.166 | -0.106 | -0.385 |
| | DeepAR | -0.142 | -0.248 | -0.157 | -0.103 | -0.360 |
| | DirectTabular | -0.168 | -0.282 | -0.184 | -0.117 | -0.410 |
| | DynamicOptimizedTheta | -0.150 | -0.255 | -0.164 | -0.105 | -0.378 |
| | NPTS | -0.404 | -0.588 | -0.422 | -0.260 | -0.982 |
| | PatchTST | -0.143 | -0.262 | -0.157 | -0.106 | -0.363 |
| | RecursiveTabular | -0.176 | -0.298 | -0.197 | -0.125 | -0.433 |
| | SeasonalNaive | -0.443 | -0.689 | -0.439 | -0.292 | -1.055 |
| | TemporalFusionTransformer | -0.143 | -0.243 | -0.161 | -0.103 | -0.364 |
| | TiDE | -0.142 | -0.251 | -0.157 | -0.105 | -0.360 |
| 51003 | **ChronosFineTuned** | **-0.138** | **-0.192** | **-0.061** | **-0.055** | **-0.452** |
| | ChronosZeroShot | -0.144 | -0.202 | -0.064 | -0.057 | -0.471 |
| | AutoETS | -0.143 | -0.202 | -0.063 | -0.057 | -0.466 |
| | DeepAR | -0.145 | -0.203 | -0.064 | -0.058 | -0.475 |
| | DirectTabular | -0.160 | -0.229 | -0.071 | -0.065 | -0.513 |

|  | model | -MAE | -RMSE | -SMAPE | -RMSLE | -MASE |
|---|---|---|---|---|---|---|
|  | DynamicOptimizedTheta | -0.145 | -0.203 | -0.064 | -0.058 | -0.470 |
|  | NPTS | -0.386 | -0.506 | -0.172 | -0.150 | -1.266 |
|  | PatchTST | -0.149 | -0.207 | -0.066 | -0.059 | -0.488 |
|  | RecursiveTabular | -0.177 | -0.252 | -0.078 | -0.071 | -0.565 |
|  | SeasonalNaive | -0.331 | -0.466 | -0.144 | -0.133 | -1.036 |
|  | TemporalFusionTransformer | -0.145 | -0.206 | -0.064 | -0.058 | -0.473 |
|  | TiDE | -0.141 | -0.197 | -0.062 | -0.056 | -0.460 |

**Table 5 Model Metrics Averaged by Forecast Horizon and Evaluation Criterion**

| predict length | model | -MAE | -RMSE | -SMAPE | -RMSLE | -MASE |
|---|---|---|---|---|---|---|
| 1 | ChronosFineTune | **-0.093** | **-0.138** | **-0.103** | **-0.054** | **-0.244** |
|  | ChronosZeroShot | -0.097 | -0.144 | -0.108 | -0.057 | -0.256 |
|  | AutoETS | -0.095 | -0.139 | -0.104 | -0.055 | -0.250 |
|  | DeepAR | -0.114 | -0.170 | -0.127 | -0.067 | -0.300 |
|  | DirectTabular | -0.113 | -0.178 | -0.126 | -0.068 | -0.287 |
|  | DynamicOptimizedTheta | -0.094 | -0.140 | -0.101 | -0.055 | -0.248 |
|  | NPTS | -0.452 | -0.623 | -0.423 | -0.252 | -1.131 |
|  | PatchTST | -0.100 | -0.161 | -0.110 | -0.061 | -0.261 |
|  | RecursiveTabular | -0.124 | -0.187 | -0.140 | -0.074 | -0.322 |
|  | SeasonalNaive | -0.403 | -0.596 | -0.378 | -0.235 | -1.010 |
|  | TemporalFusionTransformer | -0.099 | -0.151 | -0.110 | -0.060 | -0.260 |
|  | TiDE | -0.096 | -0.145 | -0.107 | -0.058 | -0.252 |
| 3 | ChronosFineTuned | **-0.115** | **-0.175** | **-0.127** | **-0.070** | **-0.304** |
|  | ChronosZeroShot | -0.121 | -0.183 | -0.135 | -0.073 | -0.321 |
|  | AutoETS | -0.123 | -0.183 | -0.137 | -0.074 | -0.326 |
|  | DeepAR | -0.127 | -0.191 | -0.140 | -0.077 | -0.336 |
|  | DirectTabular | -0.160 | -0.240 | -0.182 | -0.098 | -0.413 |
|  | DynamicOptimizedTheta | -0.121 | -0.183 | -0.134 | -0.074 | -0.321 |
|  | NPTS | -0.449 | -0.620 | -0.419 | -0.251 | -1.126 |
|  | PatchTST | -0.124 | -0.194 | -0.134 | -0.076 | -0.326 |
|  | RecursiveTabular | -0.158 | -0.239 | -0.177 | -0.096 | -0.413 |
|  | SeasonalNaive | -0.403 | -0.596 | -0.377 | -0.235 | -1.012 |
|  | TemporalFusionTransformer | -0.121 | -0.184 | -0.137 | -0.075 | -0.320 |
|  | TiDE | -0.120 | -0.184 | -0.133 | -0.074 | -0.316 |
| 6 | ChronosFineTuned | **-0.145** | **-0.227** | **-0.156** | **-0.091** | **-0.383** |
|  | ChronosZeroShot | -0.153 | -0.236 | -0.165 | -0.095 | -0.404 |
|  | AutoETS | -0.158 | -0.240 | -0.171 | -0.097 | -0.421 |
|  | DeepAR | -0.157 | -0.240 | -0.171 | -0.097 | -0.416 |
|  | DirectTabular | -0.211 | -0.317 | -0.228 | -0.128 | -0.540 |
|  | DynamicOptimizedTheta | -0.154 | -0.238 | -0.165 | -0.096 | -0.408 |
|  | NPTS | -0.447 | -0.618 | -0.416 | -0.249 | -1.121 |
|  | PatchTST | -0.152 | -0.241 | -0.162 | -0.096 | -0.401 |
|  | RecursiveTabular | -0.200 | -0.302 | -0.218 | -0.122 | -0.517 |
|  | SeasonalNaive | -0.403 | -0.592 | -0.377 | -0.234 | -1.010 |
|  | TemporalFusionTransformer | -0.152 | -0.241 | -0.164 | -0.096 | -0.401 |
|  | TiDE | -0.152 | -0.236 | -0.164 | -0.097 | -0.400 |
| 12 | ChronosFineTuned | **-0.194** | **-0.307** | **-0.200** | **-0.122** | **-0.510** |
| | ChronosZeroShot | -0.204 | -0.316 | -0.212 | -0.127 | -0.536 |
| | AutoETS | -0.218 | -0.332 | -0.225 | -0.134 | -0.578 |
| | DeepAR | -0.207 | -0.328 | -0.212 | -0.130 | -0.545 |
| | DirectTabular | -0.339 | -0.493 | -0.342 | -0.203 | -0.851 |
| | DynamicOptimizedTheta | -0.210 | -0.325 | -0.218 | -0.131 | -0.554 |
| | NPTS | -0.449 | -0.618 | -0.414 | -0.249 | -1.127 |
| | PatchTST | -0.200 | -0.315 | -0.205 | -0.125 | -0.526 |
| | RecursiveTabular | -0.252 | -0.375 | -0.270 | -0.154 | -0.646 |
| | SeasonalNaive | -0.403 | -0.591 | -0.374 | -0.232 | -1.011 |
| | TemporalFusionTransformer | -0.206 | -0.327 | -0.214 | -0.130 | -0.540 |
| | TiDE | -0.204 | -0.318 | -0.212 | -0.129 | -0.534 |
| 24 | ChronosFineTuned | **-0.256** | **-0.401** | **-0.255** | **-0.159** | **-0.666** |
| | ChronosZeroShot | -0.274 | -0.414 | -0.276 | -0.167 | -0.713 |
| | AutoETS | -0.302 | -0.442 | -0.299 | -0.183 | -0.790 |
| | DeepAR | -0.283 | -0.438 | -0.293 | -0.180 | -0.734 |
| | DirectTabular | -0.444 | -0.617 | -0.407 | -0.252 | -1.125 |
| | DynamicOptimizedTheta | -0.285 | -0.425 | -0.292 | -0.176 | -0.739 |
| | NPTS | -0.446 | -0.615 | -0.409 | -0.247 | -1.126 |
| | PatchTST | -0.264 | -0.405 | -0.263 | -0.162 | -0.687 |
| | RecursiveTabular | -0.328 | -0.481 | -0.323 | -0.198 | -0.840 |
| | SeasonalNaive | -0.401 | -0.580 | -0.377 | -0.231 | -1.007 |
| | TemporalFusionTransformer | -0.278 | -0.419 | -0.291 | -0.175 | -0.723 |
| | TiDE | -0.276 | -0.417 | -0.283 | -0.172 | -0.718 |
| 36 | ChronosFineTuned | **-0.303** | **-0.463** | **-0.295** | **-0.185** | **-0.787** |
| | ChronosZeroShot | -0.324 | -0.485 | -0.311 | -0.192 | -0.835 |
| | AutoETS | -0.378 | -0.552 | -0.356 | -0.225 | -0.989 |
| | DeepAR | -0.336 | -0.493 | -0.363 | -0.208 | -0.882 |
| | DirectTabular | -0.443 | -0.621 | -0.405 | -0.255 | -1.125 |
| | DynamicOptimizedTheta | -0.342 | -0.511 | -0.333 | -0.206 | -0.878 |
| | NPTS | -0.445 | -0.608 | -0.409 | -0.247 | -1.131 |
| | PatchTST | -0.304 | -0.461 | -0.297 | -0.183 | -0.790 |
| | RecursiveTabular | -0.361 | -0.521 | -0.369 | -0.216 | -0.915 |
| | SeasonalNaive | -0.448 | -0.650 | -0.406 | -0.253 | -1.121 |
| | TemporalFusionTransformer | -0.327 | -0.508 | -0.324 | -0.199 | -0.839 |
| | TiDE | -0.323 | -0.481 | -0.319 | -0.196 | -0.837 |
| 48 | ChronosFineTuned | **-0.330** | **-0.490** | **-0.314** | **-0.196** | **-0.840** |
| | ChronosZeroShot | -0.347 | -0.523 | -0.324 | -0.204 | -0.882 |
| | AutoETS | -0.442 | -0.638 | -0.398 | -0.256 | -1.147 |
| | DeepAR | -0.347 | -0.514 | -0.347 | -0.213 | -0.899 |
| | DirectTabular | -0.457 | -0.628 | -0.407 | -0.255 | -1.141 |
| | DynamicOptimizedTheta | -0.379 | -0.566 | -0.361 | -0.226 | -0.955 |
| | NPTS | -0.438 | -0.593 | -0.397 | -0.239 | -1.104 |
| | PatchTST | -0.328 | -0.482 | -0.310 | -0.193 | -0.841 |
| | RecursiveTabular | -0.382 | -0.549 | -0.366 | -0.226 | -0.958 |
| | SeasonalNaive | -0.453 | -0.648 | -0.409 | -0.257 | -1.135 |
| | TemporalFusionTransformer | -0.350 | -0.514 | -0.343 | -0.212 | -0.891 |
| | TiDE | -0.355 | -0.525 | -0.342 | -0.212 | -0.907 |
| 60 | ChronosFineTuned | **-0.367** | **-0.555** | **-0.335** | **-0.213** | **-0.939** |
|    | ChronosZeroShot          | -0.393 | -0.581 | -0.358 | -0.227 | -1.000 |
|    | AutoETS                  | -0.513 | -0.742 | -0.441 | -0.291 | -1.349 |
|    | DeepAR                   | -0.397 | -0.569 | -0.376 | -0.229 | -1.023 |
|    | DirectTabular            | -0.493 | -0.692 | -0.429 | -0.275 | -1.234 |
|    | DynamicOptimizedTheta    | -0.434 | -0.631 | -0.397 | -0.250 | -1.092 |
|    | NPTS                     | -0.459 | -0.629 | -0.406 | -0.248 | -1.158 |
|    | PatchTST                 | -0.366 | -0.548 | -0.340 | -0.213 | -0.932 |
|    | RecursiveTabular         | -0.436 | -0.625 | -0.391 | -0.250 | -1.090 |
|    | SeasonalNaive            | -0.508 | -0.729 | -0.443 | -0.283 | -1.256 |
|    | TemporalFusionTransformer| -0.414 | -0.604 | -0.394 | -0.246 | -1.046 |
|    | TiDE                     | -0.385 | -0.564 | -0.345 | -0.221 | -0.971 |
| 72 | ChronosFineTuned         | **-0.377** | **-0.560** | **-0.351** | **-0.221** | **-0.974** |
|    | ChronosZeroShot          | -0.400 | -0.595 | -0.361 | -0.231 | -1.028 |
|    | AutoETS                  | -0.538 | -0.759 | -0.458 | -0.306 | -1.444 |
|    | DeepAR                   | -0.401 | -0.569 | -0.372 | -0.231 | -1.037 |
|    | DirectTabular            | -0.479 | -0.665 | -0.424 | -0.268 | -1.220 |
|    | DynamicOptimizedTheta    | -0.446 | -0.647 | -0.408 | -0.260 | -1.122 |
|    | NPTS                     | -0.439 | -0.599 | -0.399 | -0.240 | -1.132 |
|    | PatchTST                 | -0.378 | -0.559 | -0.350 | -0.222 | -0.972 |
|    | RecursiveTabular         | -0.437 | -0.611 | -0.398 | -0.250 | -1.109 |
|    | SeasonalNaive            | -0.499 | -0.708 | -0.443 | -0.282 | -1.255 |
|    | TemporalFusionTransformer| -0.402 | -0.588 | -0.374 | -0.237 | -1.021 |
|    | TiDE                     | -0.399 | -0.582 | -0.357 | -0.231 | -1.013 |
| 84 | ChronosFineTuned         | **-0.394** | **-0.561** | **-0.368** | **-0.227** | **-1.019** |
|    | ChronosZeroShot          | -0.428 | -0.601 | -0.392 | -0.244 | -1.097 |
|    | AutoETS                  | -0.605 | -0.837 | -0.495 | -0.342 | -1.602 |
|    | DeepAR                   | -0.429 | -0.609 | -0.395 | -0.251 | -1.095 |
|    | DirectTabular            | -0.502 | -0.684 | -0.440 | -0.281 | -1.267 |
|    | DynamicOptimizedTheta    | -0.510 | -0.696 | -0.462 | -0.289 | -1.292 |
|    | NPTS                     | -0.444 | -0.589 | -0.410 | -0.244 | -1.143 |
|    | PatchTST                 | -0.392 | -0.550 | -0.365 | -0.225 | -1.011 |
|    | RecursiveTabular         | -0.504 | -0.691 | -0.458 | -0.287 | -1.276 |
|    | SeasonalNaive            | -0.567 | -0.774 | -0.489 | -0.313 | -1.434 |
|    | TemporalFusionTransformer| -0.423 | -0.594 | -0.392 | -0.245 | -1.084 |
|    | TiDE                     | -0.430 | -0.596 | -0.390 | -0.245 | -1.101 |
| 96 | ChronosFineTuned         | **-0.404** | **-0.586** | **-0.376** | **-0.236** | **-1.064** |
|    | ChronosZeroShot          | -0.430 | -0.648 | -0.390 | -0.256 | -1.123 |
|    | AutoETS                  | -0.603 | -0.825 | -0.493 | -0.337 | -1.650 |
|    | DeepAR                   | -0.438 | -0.615 | -0.406 | -0.253 | -1.155 |
|    | DirectTabular            | -0.515 | -0.704 | -0.453 | -0.289 | -1.321 |
|    | DynamicOptimizedTheta    | -0.503 | -0.719 | -0.475 | -0.300 | -1.304 |
|    | NPTS                     | -0.444 | -0.608 | -0.405 | -0.249 | -1.165 |
|    | PatchTST                 | -0.403 | -0.594 | -0.376 | -0.241 | -1.064 |
|    | RecursiveTabular         | -0.482 | -0.668 | -0.455 | -0.278 | -1.250 |
|    | SeasonalNaive            | -0.534 | -0.754 | -0.483 | -0.311 | -1.378 |
|    | TemporalFusionTransformer| -0.412 | -0.607 | -0.377 | -0.245 | -1.072 |
|    | TiDE                     | -0.425 | -0.623 | -0.385 | -0.252 | -1.107 |
| 108| ChronosFineTuned         | **-0.414** | **-0.603** | **-0.382** | **-0.241** | **-1.066** |

|     | Model | | | | | |
| --- | --- | --- | --- | --- | --- | --- |
|     | ChronosZeroShot | -0.457 | -0.639 | -0.413 | -0.256 | -1.154 |
|     | AutoETS | -0.646 | -0.867 | -0.510 | -0.345 | -1.724 |
|     | DeepAR | -0.483 | -0.688 | -0.433 | -0.278 | -1.246 |
|     | DirectTabular | -0.502 | -0.690 | -0.441 | -0.279 | -1.262 |
|     | DynamicOptimizedTheta | -0.507 | -0.703 | -0.475 | -0.287 | -1.272 |
|     | NPTS | -0.455 | -0.613 | -0.413 | -0.247 | -1.157 |
|     | PatchTST | -0.421 | -0.613 | -0.388 | -0.244 | -1.074 |
|     | RecursiveTabular | -0.535 | -0.717 | -0.555 | -0.303 | -1.351 |
|     | SeasonalNaive | -0.545 | -0.748 | -0.484 | -0.303 | -1.358 |
|     | TemporalFusionTransformer | -0.450 | -0.638 | -0.406 | -0.256 | -1.121 |
|     | TiDE | -0.442 | -0.628 | -0.394 | -0.249 | -1.105 |
| 120 | **ChronosFineTuned** | **-0.433** | **-0.619** | **-0.391** | **-0.254** | **-1.142** |
|     | ChronosZeroShot | -0.455 | -0.631 | -0.412 | -0.259 | -1.196 |
|     | AutoETS | -0.685 | -0.914 | -0.527 | -0.367 | -1.873 |
|     | DeepAR | -0.431 | -0.610 | -0.402 | -0.250 | -1.130 |
|     | DirectTabular | -0.534 | -0.720 | -0.461 | -0.296 | -1.370 |
|     | DynamicOptimizedTheta | -0.499 | -0.690 | -0.432 | -0.280 | -1.283 |
|     | NPTS | -0.472 | -0.628 | -0.417 | -0.257 | -1.225 |
|     | PatchTST | -0.439 | -0.612 | -0.400 | -0.250 | -1.151 |
|     | RecursiveTabular | -0.541 | -0.739 | -0.477 | -0.311 | -1.405 |
|     | SeasonalNaive | -0.549 | -0.747 | -0.466 | -0.300 | -1.394 |
|     | TemporalFusionTransformer | -0.458 | -0.642 | -0.410 | -0.260 | -1.183 |

# 7 Conflict of Interest

*The authors declare that the research was conducted in the absence of any commercial or financial relationships that could be construed as a potential conflict of interest.*

# 8 Author Contributions

Yilin Zhai was responsible for the overall conceptualization, data analysis, experimental design, and drafting of the manuscript; Hongyuan Shi participated in research discussions, provided valuable guidance and revisions, and served as the corresponding author; Chao Zhan reviewed the overall direction of the paper and offered constructive comments; Qing Wang carefully reviewed the manuscript and provided feedback and suggestions; Zaijin You assisted with data discussions and contributed to several sections of the manuscript; and Nan Wang participated in reviewing and revising the final draft. All authors have read and approved the final version of the manuscript.

# 9 Funding

*The author(s) declare financial support was received for the research, authorship, and/or publication of this article. This research was supported by the National Key R&D Program of China (2023YFC3007900, 2023YFC3007905), the National Natural Science Foundation Key Project (42330406, 42476163), and the Yantai Science and Technology Innovation Project (2023JCYJ094).*

# 10 Data Availability Statement

The datasets [ANALYZED] for this study can be found in the National Data Buoy Center (NDBC) repository [LINK: https://www.ndbc.noaa.gov/].

# 11  Reference

**1**      Alfredo, C. S., and Adytia, D. A. (2022). Time Series Forecasting of Significant Wave Height using GRU, CNN-GRU, and LSTM. *Jurnal RESTI (Rekayasa Sistem dan Teknologi Informasi)* 6, 776–781. doi: 10.29207/resti.v6i5.4160

Ansari, A. F., Stella, L., Turkmen, C., Zhang, X., Mercado, P., Shen, H., et al. (2024). Chronos: Learning the Language of Time Series. doi: 10.48550/arXiv.2403.07815

Booij, N., Ris, R. C., and Holthuijsen, L. H. (1999). A third-generation wave model for coastal regions: 1. Model description and validation. *Journal of Geophysical Research: Oceans* 104, 7649–7666. doi: 10.1029/98JC02622

Cho, K., Merrienboer, B. van, Gulcehre, C., Bahdanau, D., Bougares, F., Schwenk, H., et al. (2014). Learning Phrase Representations using RNN Encoder-Decoder for Statistical Machine Translation. doi: 10.48550/arXiv.1406.1078

Das, A., Kong, W., Leach, A., Mathur, S., Sen, R., and Yu, R. (2024). Long-term Forecasting with TiDE: Time-series Dense Encoder. doi: 10.48550/arXiv.2304.08424

Duong, N.-H., Nguyen, M.-T., Nguyen, T.-H., and Tran, T.-P. (2023). Application of Seasonal Trend Decomposition using Loess and Long Short-Term Memory in Peak Load Forecasting Model in Tien Giang. *Engineering, Technology & Applied Science Research* 13, 11628–11634. doi: 10.48084/etasr.6181

Erickson, N., Mueller, J., Shirkov, A., Zhang, H., Larroy, P., Li, M., et al. (2020). AutoGluon-Tabular: Robust and Accurate AutoML for Structured Data. doi: 10.48550/arXiv.2003.06505

Fan, S., Xiao, N., and Dong, S. (2020). A novel model to predict significant wave height based on long short-term memory network. *Ocean Engineering* 205, 107298. doi: 10.1016/j.oceaneng.2020.107298

Fiorucci, J. A., Pellegrini, T. R., Louzada, F., Petropoulos, F., and Koehler, A. B. (2016). Models for optimising the theta method and their relationship to state space models. *International Journal of Forecasting* 32, 1151–1161. doi: 10.1016/j.ijforecast.2016.02.005

Gao, R., Li, R., Hu, M., Suganthan, P. N., and Yuen, K. F. (2023). Dynamic ensemble deep echo state network for significant wave height forecasting. *Applied Energy* 329, 120261. doi: 10.1016/j.apenergy.2022.120261

Gorishniy, Y., Rubachev, I., Khrulkov, V., and Babenko, A. (2023). Revisiting Deep Learning Models for Tabular Data. doi: 10.48550/arXiv.2106.11959

Group, T. W. (1988). The WAM Model—A Third Generation Ocean Wave Prediction Model. Available at: https://journals.ametsoc.org/view/journals/phoc/18/12/1520-0485_1988_018_1775_twmtgo_2_0_co_2.xml (Accessed February 12, 2025).

Huang, X., Tang, J., and Shen, Y. (2024). Long time series of ocean wave prediction based on PatchTST model. *Ocean Engineering* 301, 117572. doi: 10.1016/j.oceaneng.2024.117572

Hyndman, R.J., Athanasopoulos, G., 2021. Forecasting: Principles and Practice, 3rd ed. OTexts, Melbourne, Australia. Available at: https://otexts.com/fpp3/ (accessed 20 February 2025).

Karita, S., Chen, N., Hayashi, T., Hori, T., Inaguma, H., Jiang, Z., et al. (2019). A Comparative Study on Transformer vs RNN in Speech Applications., in *2019 IEEE Automatic Speech Recognition and Understanding Workshop (ASRU)*, 449–456. doi: 10.1109/ASRU46091.2019.9003750

Kaur, M., and Mohta, A. (2019). A Review of Deep Learning with Recurrent Neural Network., in *2019 International Conference on Smart Systems and Inventive Technology (ICSSIT)*, 460–465. doi: 10.1109/ICSSIT46314.2019.8987837

Kumar, N. K., Savitha, R., and Mamun, A. A. (2017). Regional ocean wave height prediction using sequential learning neural networks. *Ocean Engineering* 129, 605–612. doi: 10.1016/j.oceaneng.2016.10.033

Li, G., Weiss, G., Mueller, M., Townley, S., and Belmont, M. R. (2012). Wave energy converter control by wave prediction and dynamic programming. *Renewable Energy* 48, 392–403. doi: 10.1016/j.renene.2012.05.003

Liang, B., Gao, H., and Shao, Z. (2019). Characteristics of global waves based on the third-generation wave model SWAN. *Marine Structures* 64, 35–53. doi: 10.1016/j.marstruc.2018.10.011

Lim, B., Arik, S. O., Loeff, N., and Pfister, T. (2020). Temporal Fusion Transformers for Interpretable Multi-horizon Time Series Forecasting. doi: 10.48550/arXiv.1912.09363

Liu, J., Bian, Y., Lawson, K., and Shen, C. (2024a). Probing the limit of hydrologic predictability with the Transformer network. *Journal of Hydrology* 637, 131389. doi: 10.1016/j.jhydrol.2024.131389

Liu, Y., Lu, W., Wang, D., Lai, Z., Ying, C., Li, X., et al. (2024b). Spatiotemporal wave forecast with transformer-based network: A case study for the northwestern Pacific Ocean. *Ocean Modelling* 188, 102323. doi: 10.1016/j.ocemod.2024.102323

Nie, Y., Nguyen, N. H., Sinthong, P., and Kalagnanam, J. (2023). A Time Series is Worth 64 Words: Long-term Forecasting with Transformers. doi: 10.48550/arXiv.2211.14730

Oh, J., and Suh, K.-D. (2018). Real-time forecasting of wave heights using EOF – wavelet – neural network hybrid model. *Ocean Engineering* 150, 48–59. doi: 10.1016/j.oceaneng.2017.12.044

Pourpanah, F., Abdar, M., Luo, Y., Zhou, X., Wang, R., Lim, C. P., et al. (2023). A Review of Generalized Zero-Shot Learning Methods. *IEEE Transactions on Pattern Analysis and Machine Intelligence* 45, 4051–4070. doi: 10.1109/TPAMI.2022.3191696

Reichstein, M., Camps-Valls, G., Stevens, B., Jung, M., Denzler, J., Carvalhais, N., et al. (2019). Deep learning and process understanding for data-driven Earth system science. *Nature* 566, 195–204. doi: 10.1038/s41586-019-0912-1

Rumelhart, D. E., Hinton, G. E., and Williams, R. J. (1986). Learning representations by back-propagating errors. *Nature* 323, 533–536. doi: 10.1038/323533a0

Salinas, D., Flunkert, V., and Gasthaus, J. (2019). DeepAR: Probabilistic Forecasting with Autoregressive Recurrent Networks. doi: 10.48550/arXiv.1704.04110

Tolman, H. L. (n.d.). User manual and system documentation of WAVEWATCH III TM version 3.14.

Varela, D. A. B., Ongsakul, W., and Benitez, I. B. (2024). Machine Learning Applications in Wave Energy Forecasting., in *2024 International Conference on Sustainable Energy: Energy Transition and Net-Zero Climate Future (ICUE)*, 1–8. doi: 10.1109/ICUE63019.2024.10795514

Vaswani, A., Shazeer, N., Parmar, N., Uszkoreit, J., Jones, L., Gomez, A. N., et al. (2023). Attention Is All You Need. doi: 10.48550/arXiv.1706.03762

Wang, L., Wang, X., Dong, C., and Sun, Y. (2024). Wave predictor models for medium and long term based on dual attention-enhanced Transformer. *Ocean Engineering* 310, 118761. doi: 10.1016/j.oceaneng.2024.118761

Wu, X., Zhang, D., Guo, C., He, C., Yang, B., and Jensen, C. S. (2021). AutoCTS: Automated Correlated Time Series Forecasting -- Extended Version. doi: 10.48550/arXiv.2112.11174

Yang, S., Deng, Z., Li, X., Zheng, C., Xi, L., Zhuang, J., et al. (2021). A novel hybrid model based on STL decomposition and one-dimensional convolutional neural networks with positional encoding for significant wave height forecast. *Renewable Energy* 173, 531–543. doi: 10.1016/j.renene.2021.04.010

Yevnin, Y., Chorev, S., Dukan, I., and Toledo, Y. (2023). Short-term wave forecasts using gated recurrent unit model. *Ocean Engineering* 268, 113389. doi: 10.1016/j.oceaneng.2022.113389

Zhang, J., Zhao, X., Jin, S., and Greaves, D. (2022). Phase-resolved real-time ocean wave prediction with quantified uncertainty based on variational Bayesian machine learning. *Applied Energy* 324, 119711. doi: 10.1016/j.apenergy.2022.119711

Zhang, M., Yuan, Z.-M., Dai, S.-S., Chen, M.-L., and Incecik, A. (2024). LSTM RNN-based excitation force prediction for the real-time control of wave energy converters. *Ocean Engineering* 306, 118023. doi: 10.1016/j.oceaneng.2024.118023

S. Hochreiter and J. Schmidhuber, "Long Short-Term Memory," in *Neural Computation*, vol. 9, no. 8, pp. 1735-1780, 15 Nov. 1997, doi: 10.1162/neco.1997.9.8.1735.